\documentclass[11pt]{article}

\usepackage[final]{acl}
\usepackage{amsmath} 
\usepackage{amssymb}
\usepackage{booktabs}
\usepackage{adjustbox}
\usepackage{enumitem}
\usepackage{times}
\usepackage{latexsym}
\usepackage{cleveref}
\usepackage{multirow}
\usepackage{algorithm}
\usepackage{cancel}
\usepackage{tikz}
\usepackage{hyperref}

\newcommand{\circled}[1]{\tikz[baseline=(char.base)]{\node[shape=circle,draw,inner sep=1pt] (char) {#1};}}
\usepackage{algpseudocode}
\usepackage[T1]{fontenc}

\usepackage[utf8]{inputenc}

\usepackage{microtype}

\usepackage{inconsolata}

\usepackage{graphicx}

\newif\ifcomments
\commentstrue  

\title{From Local to Global: Revisiting Structured Pruning Paradigms for Large Language Models}

\author{
 \textbf{Ziyan Wang\textsuperscript{1}},
 \textbf{Enmao Diao\textsuperscript{2}},
 \textbf{Qi Le\textsuperscript{3}},
\\
 \textbf{Pu Wang\textsuperscript{1}},
 \textbf{Minwoo Lee\textsuperscript{1}},
 \textbf{Shu-ping Yeh\textsuperscript{4}},
\\
 \textbf{Evgeny V Stupachenko\textsuperscript{4}},
 \textbf{Hao Feng\textsuperscript{4}},
 \textbf{Li Yang\textsuperscript{1}},
\\
 \textsuperscript{1}University of North Carolina at Charlotte,
 \textsuperscript{2}DreamSoul,
\textsuperscript{3}University of Minnesota,
 \textsuperscript{4}Intel Corporation,
\\
 \texttt{zwang53@charlotte.edu} \quad \texttt{lyang50@charlotte.edu}
 \\
}

\begin{document}
\maketitle

\begin{abstract}
Structured pruning is a practical approach to deploying large language models (LLMs) efficiently, as it yields compact, hardware-friendly architectures. However, the dominant local paradigm is task-agnostic: by optimizing layer-wise reconstruction rather than task objectives, it tends to preserve perplexity or generic zero-shot behavior but fails to capitalize on modest task-specific calibration signals, often yielding limited downstream gains. We revisit global structured pruning and present GISP, \textit{Global Iterative Structured Pruning}, a post-training method that removes attention heads and MLP channels using first-order, loss-based important scores aggregated at the structure level with block-wise normalization. Built on this global importance metric, GISP adopts
an iterative schedule, rather than one-shot pruning, stabilizes accuracy at higher sparsity, and mitigates perplexity collapse without requiring intermediate fine-tuning. Importantly, the iterative pruning forms nested subnetworks that support a ``prune-once, deploy-many'' workflow. Furthermore, GISP defines structural importance directly with respect to a target loss, making it easy to adapt pruning to task-specific objectives. In this work, we use perplexity for language modeling and a margin-based objective for decision-style tasks. Extensive experiments show that across Llama2‑7B/13B, Llama3‑8B, and Mistral‑0.3‑7B, GISP consistently lowers WikiText‑2 perplexity and improves downstream accuracy, with especially strong gains at 40–50\% sparsity; on DeepSeek-R1-Distill-Llama-3-8B and Qwen3-8B with GSM8K, task‑aligned calibration substantially boosts exact‑match accuracy. The implementation is available at \href{https://github.com/uncc-efficient-ai/GISP}{the official GISP repository}.

\end{abstract}

\section{Introduction}
Pruning~\cite {ma2023llm, frantar2023sparsegptmassivelanguagemodels, sun2024simpleeffectivepruningapproach, kim2024shortenedllamadepthpruning,an2023fluctuationbasedadaptivestructuredpruning} is a fundamental technique for compressing neural networks by removing redundant parameters while preserving accuracy. Broadly, existing approaches fall into two categories: unstructured pruning, which removes element-wise weights without shrinking the model architecture, and structured pruning, which eliminates entire groups of weights (e.g., channels, attention heads, layers). It is well established that unstructured pruning can achieve higher sparsity levels but typically requires specialized sparse computation kernels to realize runtime speedups, whereas structured pruning inherently produces compact, hardware-friendly architectures and is therefore preferred for practical deployment.
With the rapid emergence of Large Language Models (LLMs)~\cite{touvron2023llama,touvron2023llama2openfoundation,openai2024gpt4technicalreport,vicuna2023,workshop2023bloom176bparameteropenaccessmultilingual,grattafiori2024llama3herdmodels} containing billions of parameters, pruning has become critical to improve inference efficiency and to enable deployment on resource-constrained devices.

\begin{figure}[t]
\begin{center}
\centerline{\includegraphics[width=1.05\columnwidth]{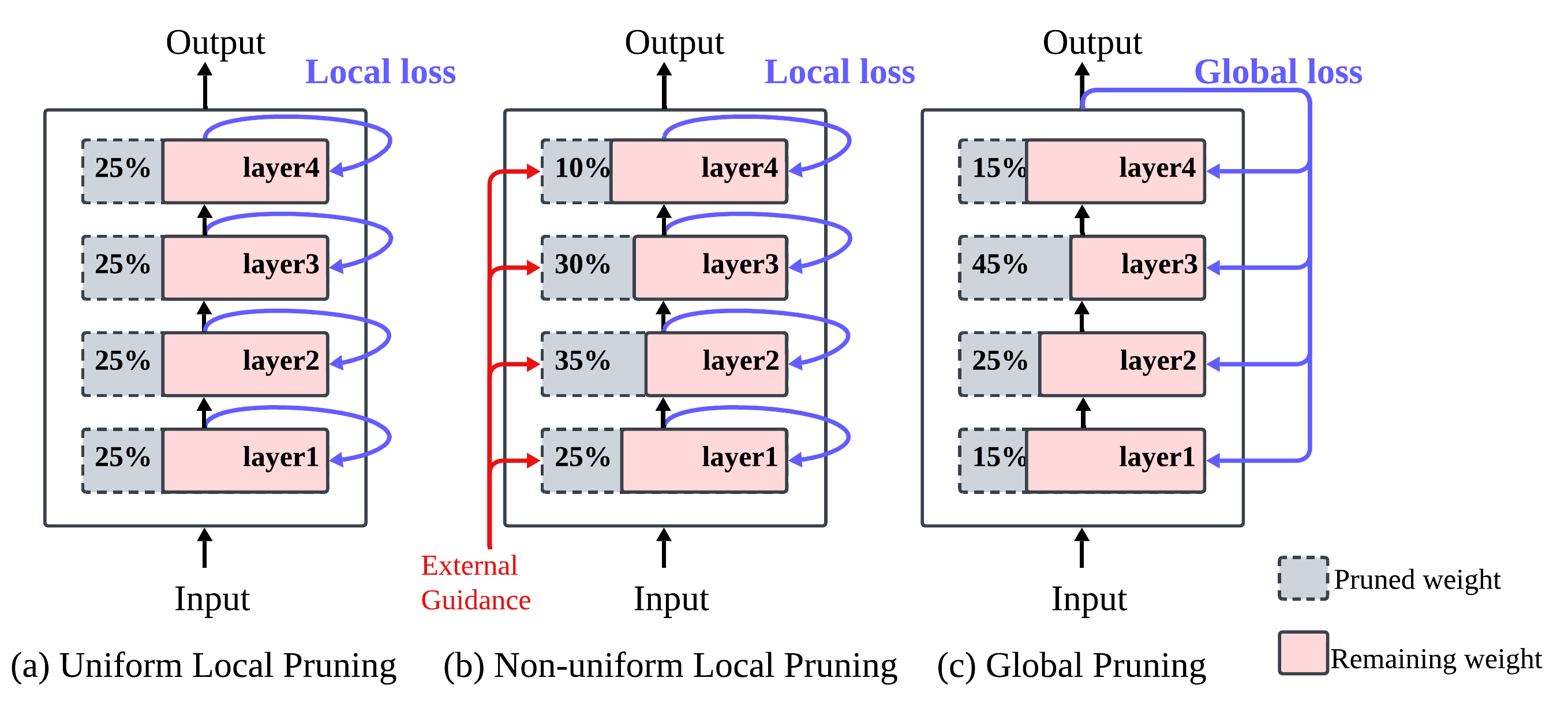}}
\vspace{-2mm}
\caption{Comparison between (a) local pruning, which uses layer-wise reconstruction loss as the importance criterion, (b) non-uniform variants of local pruning, and (c) global pruning, which directly considers the impact of weight pruning on the final model loss.}
\label{fig:front}
\vspace{-3em}
\end{center}
\end{figure}

The dominant paradigm for pruning LLMs is \textbf{local pruning}, exemplified by methods like SparseGPT~\cite{frantar2023sparsegptmassivelanguagemodels} and Wanda~\cite{sun2024simpleeffectivepruningapproach}. These methods gained significant attention due to their simplicity and efficiency, breaking down the model-wide optimization into layer-wise sub-problems (Fig.\ref{fig:front}(a)). This decomposition allows them to prune each layer gradually, typically by minimizing a layer-wise reconstruction with calibration data, offering a post-training solution. Furthermore, to mitigate the rigidity of uniform sparsity, recent work explores \textbf{non-uniform local pruning} (Fig.~\ref{fig:front}(b)), which adjusts layer-wise ratios based on estimated importance.
Methods such as OWL~\cite{yin2025outlierweighedlayerwisesparsity}, FLAP~\cite{an2023fluctuationbasedadaptivestructuredpruning}, and DarwinLM~\cite{tang2025darwinlmevolutionarystructuredpruning} leverage activation statistics or evolutionary search to assign non-uniform sparsity.
While these approaches improve accuracy, they remain rooted in layer-wise reconstruction and introduce notable algorithmic complexity and overhead.

While local structured pruning is efficient and preserves broad behavior (often reflected in perplexity or generic zero-shot scores), \textbf{its objective is task-agnostic}. When modest task-informed calibration is available, local methods rarely capitalize on it, yielding limited downstream gains. \emph{This gap calls for a loss-aligned alternative defined at the model level, instead of a local proxy.} We therefore revisit \textbf{global pruning} (Fig.\ref{fig:front}(c)) for LLMs and develop \textbf{GISP—Global Iterative Structured Pruning}. Unlike local approaches that optimize layer reconstructions, global pruning defines importance with respect to a model-level loss, naturally inducing non-uniform sparsity without extra heuristics. Operationally, GISP aggregates first‑order, loss‑based importance at the structure level (attention heads and MLP channels) with block‑wise normalization, and we study it in a post‑training setting (no fine‑tuning between steps) to match practical constraints.



Building on this formulation, we first validate that making global pruning iterative fundamentally changes its behavior. A gradual, ratio-scheduled process turns the otherwise unstable one-shot global pruning into a robust procedure that preserves model quality even at high sparsity. Furthermore, the same iterative trajectory also reveals a nested structure across sparsity levels, showing that iterative global pruning can serve as a single, continuous optimization rather than a series of independent runs, thereby enabling a ``prune-once, deploy-many'' workflow. Finally, because importance is defined by a model-level loss, GISP can directly integrate task objectives, bridging the gap between generic compression and task-aware optimization; this property consistently yields stronger downstream accuracy across models and pruning ratios.

We summarize our contributions as follows:

\begin{itemize}
\item We present GISP, a simple and effective global iterative structured pruning framework for LLMs that operates post‑training and stabilizes performance at high sparsity.
\item We demonstrate that iterative global pruning follows a smooth, nested trajectory of subnetworks, enabling a ``prune-once, deploy-many'' workflow with a competitive amortized time cost per usable model compared to local pruning baselines. 
\item We examine the task-specific property of GISP and instantiate task‑specific global importance via multiple objectives. Extensive experiments demonstrate consistent downstream gains across models and pruning ratio levels.
\end{itemize}

\section{Preliminary: Local \textit{vs} Global Pruning}
\label{sec:preliminary}
\subsection{Local Pruning and Non-uniform Variant}

\paragraph{Rationale of local pruning.} Given a pre-trained model with weights as \(\theta\) and a set of calibration dataset 
\(D=\{ (x_i, y_i) \}_{i=1}^N\) with  \(N\) samples, the structure pruning in LLMs with $L$ transformer~\cite{vaswani2023attentionneed} layers can be interpreted as finding optimal \(\widehat{\theta}\) under desired sparsity constraints by removing 
sets of coupled structures \(W_{l}\) from \(G = (\{ W_{l, \text{Attn}} \}_{l=1}^L, \{ W_{l,\text{MLP}} \}_{l=1}^L)\) with minimal 
error
on a pre-defined objective function.

As introduced by the pioneering work OBS~\cite{NIPS1992_303ed4c6} and layer-wise OBS ~\cite{dong2017learningprunedeepneural}, local pruning defines the objective function by breaking down the problem of full model compression into sub-problems for each layer. It constructs a local loss to measure the $L_2$ error between the outputs of the unpruned and pruned layers, which can be formulated as:
\begin{equation}
\small
    \min_{M_{\ell}, \widehat{W}_{\ell}}
\Bigl\|
W_{\ell} \, X_{\ell}
-
\bigl(M_{\ell} \odot \widehat{W}_{\ell}\bigr) \, X_{\ell}
\Bigr\|_{2}^{2},
\label{eqt:l2_local}
\end{equation}
where \(W_{\ell}\) is the original weight of layer \(\ell\), \(X_{\ell}\) is the input to layer \(\ell\), \(M_{\ell}\) is the binary mask indicating which weights to keep, and \(\widehat{W}_{\ell}\) \textcolor{black}{denotes the post-pruning weights, which may be kept fixed as \(W_{\ell}\) or re-optimized given \(M_{\ell}\)}.

Among local structured pruning methods, SparseGPT~\cite{frantar2023sparsegptmassivelanguagemodels} formulates pruning as a sparse regression problem solved via an approximate Hessian inversion.
ZipLM~\cite{kurtic2023ziplminferenceawarestructuredpruning} extends the OBS formulation to structured pruning and performs inference-aware search over structures.
Wanda~\cite{sun2024simpleeffectivepruningapproach} simplifies SparseGPT’s importance to weight–activation products, achieving similar accuracy with higher efficiency.
LLM-Pruner~\cite{ma2023llm} further prunes entire attention heads and MLP channels using gradient information to capture inter-structure dependencies. \textcolor{black}{GBLM~\cite{das2024sizegradientsshapepruning} augments Wanda's scores with gradient magnitude, while Wanda++~\cite{yang2025wandapruninglargelanguage} adds block-level gradients from activation reconstruction to improve layer-wise pruning; Pruner-Zero~\cite{dong2024prunerzeroevolvingsymbolicpruning} instead uses genetic programming to discover gradient-based layer-local importance metrics from weight/activation/gradient signals.} Finally, several works explore layer-wise pruning~\cite{kim2024shortenedllamadepthpruning,men2024shortgptlayerslargelanguage}, such as ShortGPT~\cite{men2024shortgptlayerslargelanguage}, which leverages layer-wise activation similarity. 

\noindent\textbf{Non-uniform variants of local pruning.}
To overcome the limitation of uniform sparsity in layer-wise pruning, several works introduce non-uniform local pruning (Fig.~\ref{fig:front}(b)) that adjusts pruning ratios across layers based on estimated importance.
These methods extend the layer-wise reconstruction paradigm by incorporating inter-layer sensitivity through diverse heuristics:
FLAP~\cite{an2023fluctuationbasedadaptivestructuredpruning} exploits activation variability to assign flexible sparsity,
OWL~\cite{yin2025outlierweighedlayerwisesparsity} reweights layers according to outlier statistics in activations,
and DarwinLM~\cite{tang2025darwinlmevolutionarystructuredpruning} performs a training-aware evolutionary search to identify optimal sparsity configurations.

\subsection{Global Pruning}

Global pruning aims to find a global sparsity mask \(M\) and possibly updated weights \(\widehat{W}\) to minimize the global loss between the final outputs of the uncompressed and compressed model. 
Hence, the learning objective can be formulated as:
\begin{equation}
\small
\min_{M,\ \widehat{W}} \,
\Delta\mathcal{L}\Bigl(f\bigl(X;\,M \odot \widehat{W}\bigr), \; f(X;W)\Bigr),
\label{eqt:global_pruning}
\end{equation}
where \(f\) is the forward function, \(X\) denotes the inputs, \(W\) is the original (pre-trained) weight, \(M\) is the binary mask indicating which weights remain. Following the idea from OBD~\cite{NIPS1989_6c9882bb} of conducting the Taylor series towards loss distance on parameter perturbation caused by pruning, we have element-wise importance given by
\begin{equation}
\small
\begin{aligned}
I_{W_i^j} 
= \bigl|\Delta \mathcal{L}(D)\bigr|
= \bigl|\mathcal{L}(D;\theta_{W_i^j}) - \mathcal{L}(D;\theta_{W_i^j=0}) \bigr|
= \\
\left|\,
\frac{\partial \mathcal{L}(D)}{\partial W_i^j}\,W_i^j
\;-\;\frac{1}{2}\,W_i^j H_{jj}\,W_i^j
\;+\;\mathcal{O}\bigl(\|W_i^j\|^3\bigr)
\right|,
\end{aligned}
\label{eq:score_glocal}
\end{equation}
where $I_{W_i^j}$ marks the $j-{th}$ estimated importance of element in $\theta$, $H_{jj}$ is diagonal of the hessian matrix. Global pruning has been extensively studied in smaller networks such as CNNs, Vision Transformers, and compact language models~\cite{molchanov2016pruning,yang2023globalvisiontransformerpruning,diao2023pruningdeepneuralnetworks}, consistently outperforming local approaches~\cite{blalock2020state,diao2023pruningdeepneuralnetworks}.
In LLMs, \textcolor{black}{SparseLLM \cite{bai2024sparsellmglobalpruningpretrained} approximates Eq.~\ref{eqt:global_pruning} through a sequence of block-level reconstruction
constraints with iterative hidden-state updates to propagate pruning effects across blocks.} LLM-Pruner~\cite{ma2023llm} applies Eq.~\ref{eq:score_glocal} for element-wise importance and explores structure-level aggregation, while LoRAPrune~\cite{zhang2024loraprunestructuredpruningmeets} adapts it to LoRA for memory-efficient fine-tuning. Although higher-order derivatives can be included, prior work in both CNNs and LLMs~\cite{ma2023llm,molchanov2019importanceestimationneuralnetwork,zhang2024loraprunestructuredpruningmeets} shows that first-order information alone is sufficient for competitive results.


\noindent\textbf{Motivation.}
While local structured pruning is appealing for its efficiency, it remains fundamentally task-agnostic.
These methods minimize layer-wise reconstruction errors to preserve input–output similarity with the dense model, which maintains perplexity and generic zero-shot behavior but does not optimize downstream accuracy.
As shown in Table \ref{tab:local_pruning_downstream}, we evaluate local methods on CMQA using Llama 2-13B under two calibration settings: a generic C4 corpus and task-specific CMQA samples.
Even with task-informed calibration, the improvement is marginal, indicating that local pruning cannot effectively exploit task signals.

In contrast, global pruning defines importance with respect to the overall model loss, naturally producing non-uniform sparsity patterns.
Because its importance scores are computed on calibration data, global pruning can directly align pruning decisions with downstream objectives.
This motivates our exploration of task-specific global iterative structured pruning, which unifies the efficiency of post-training methods with the flexibility to incorporate task-aware objectives.

\section{Method}

\subsection{A Naive Case Study: One-shot Global Pruning}\label{sec:naive-one-shot-gp}


To assess the effectiveness and limitations of global pruning, we first replicate prior pruning protocols designed for smaller models~\cite{frankle2018lottery,mallya2018packnetaddingmultipletasks}. Specifically, we conduct the following procedure: 
\color{black}
\begin{enumerate}[topsep=2pt, itemsep=1pt, parsep=0pt, partopsep=0pt]
    \item Given a calibration dataset \(D = \{ (x_i, y_i) \}_{i=1}^N\), we calculate the importance score of each weight element by using the first order term from \cref{eq:score_glocal}.
    \item We construct the structure-wise importance scores by summing a group of importance scores of each weight element, where a group is defined as one head of the attention layer or a channel of the MLP layer. After the aggregation, a normalization process is conducted in consideration of group size among different structures, allowing all structural importance to be comparable across the model. 
    \item Given a current pruning ratio $\rho$, globally rank the normalized structure-wise scores and select the Top-K lowest-scoring structured weights for pruning.
\end{enumerate}
\color{black}
\begin{figure}[t]
\begin{center}
\centerline{\includegraphics[width=1.0\columnwidth]{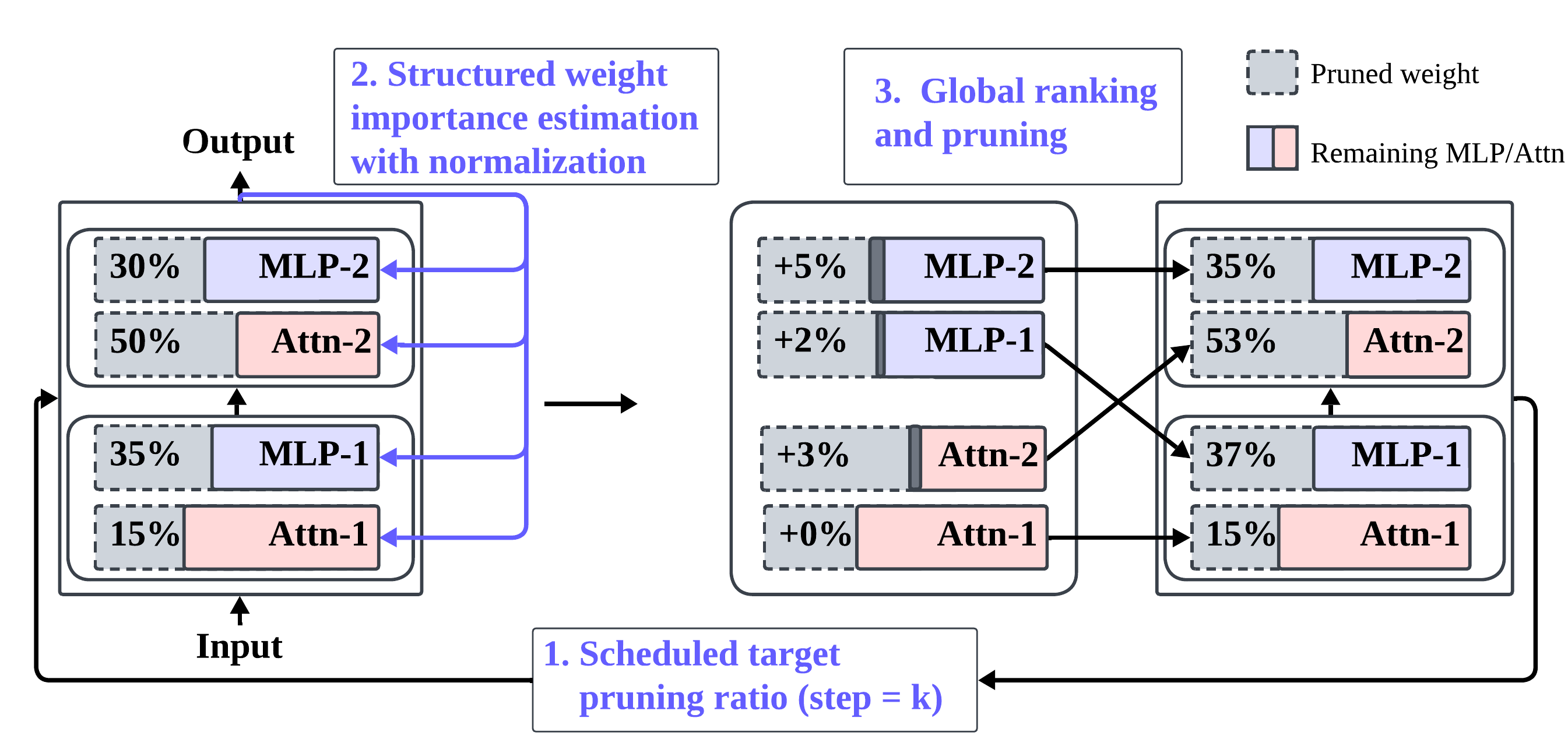}}
\vspace{-2mm}
\caption{A detailed overview of the GISP. GISP performs iterative structured pruning guided by a ratio scheduler. 
}
\vspace{-2mm}
\label{overview-fig}
\vspace{-2em}
\end{center}
\end{figure}

Compared to prior pruning works on small models~\cite{frankle2018lottery,mallya2018packnetaddingmultipletasks}, we make two changes: (1) these methods typically adopt a prune–then–fine-tune paradigm. In contrast, we evaluate performance in the setting of post-training pruning without fine-tuning, consistent with common local pruning practices for LLMs, as the computational and memory costs of fine-tuning after each pruning iteration are prohibitive, and (2) we observe that attention blocks exhibit substantially higher importance scores than MLP blocks (see Fig \ref{fig:ppl_its_analysis}(c)), often by an order of magnitude. To address this imbalance, we normalize importance scores within attention and MLP blocks separately, which empirically yields improved accuracy.

\noindent\textbf{Empirical study.} We perform experiments on one-shot global pruning and compare it with one representative structured local pruning method, Wanda. These experiments are conducted on Llama2-7B with a target pruning ratio of 20-50\%. As shown in Table \ref{tab:oneshotGP}, one-shot global pruning surpasses Wanda at low ratios but degrades at high sparsity, indicating that naive one-shot pruning is viable yet unstable, especially in high-pruning-ratio regions, which motivates our iterative strategy introduced next.

\begin{table}
  \centering
  \caption{Overall average CMQA accuracy (\%) under different calibration datasets and pruning ratios. Local pruning methods have limited performance improvement, even with a task-informed calibration dataset.}
  \vspace{-2mm}
  \label{tab:local_pruning_downstream}
  \resizebox{0.9\linewidth}{!}{
  \begin{tabular}{llcc}
    \toprule
    \textbf{Method}        & \textbf{Calibration Data}  & \textbf{Pruning Ratio} & \textbf{AVG ACC}   \\
    \midrule
    \multirow{4}{*}{Wanda-sp} 
                 & \multirow{2}{*}{C4}   & 20\%     & 66.36 \\
                 &                       & 40\%     & 58.24 \\
    \cmidrule(l){2-4}
                 & \multirow{2}{*}{CMQA} & 20\%     & 66.30 \\
                 &                       & 40\%     & 59.11 \\
    \midrule
    \multirow{4}{*}{LLM-Pruner}
                 & \multirow{2}{*}{C4}   & 20\%     & 65.63 \\
                 &                       & 40\%     & 55.06 \\
    \cmidrule(l){2-4}
                 & \multirow{2}{*}{CMQA} & 20\%     & 57.30 \\
                 &                       & 40\%     & 41.99 \\
    \midrule
    \multirow{4}{*}{FLAP}
                 & \multirow{2}{*}{C4}   & 20\%     & 66.15 \\
                 &                       & 40\%     & 61.73 \\
    \cmidrule(l){2-4}
                 & \multirow{2}{*}{CMQA} & 20\%     & 65.00 \\
                 &                       & 40\%     & 59.89 \\
    \midrule
    \multirow{4}{*}{OWL}
                 & \multirow{2}{*}{C4}   & 20\%     & 66.62 \\
                 &                       & 40\%     & 59.87 \\
    \cmidrule(l){2-4}
                 & \multirow{2}{*}{CMQA} & 20\%     & 66.98 \\
                 &                       & 40\%     & 60.84 \\
    \bottomrule
  \end{tabular}
  }
    \vspace{-2mm}
\end{table}

\subsection{GISP: Global Iterative Structured Pruning}

\subsubsection{Stabilizing High-Ratio Pruning}
\paragraph{Motivation.}
We hypothesize that one potential issue of one-shot global pruning is that it removes a large portion of weights at once, increasing the risk of over-pruning important weights.  A potential solution to this issue is iterative global pruning, which gradually prunes the model by applying a small pruning ratio in each step. This approach enables more precise identification of truly redundant weights, leveraging iterative feedback to refine pruning decisions. 

\begin{table}[t]
\caption{Evaluation of one-shot global pruning (marked as one-shot GP) on perplexity (PPL) with C4 as the calibration dataset.}
\centering
\label{tab:oneshotGP}
  \vspace{-2mm}
\begin{adjustbox}{width=0.35\textwidth}
\begin{tabular}{lccc}
\toprule
\textbf{Method} & \textbf{Pruning ratio} & \textbf{Wiki2} & \textbf{PTB} \\
\midrule
    \multirow{4}{*}{Wanda-sp}
  & 20\% &  22.71 & 101.23 \\
  & 30\% &  35.43 & 138.41 \\
  & 40\% &  51.85 & 185.09 \\
  & 50\% &  81.47 & 218.31 \\
  \midrule
  \multirow{4}{*}{One-shot GP}
    & 20\% &  17.93	 & 63.09  \\
    & 30\% &  26.99	 & 81.48  \\
    & 40\% &  53.45	 & 151.80  \\
    & 50\% & 159.47	 & 353.55  \\
  \bottomrule
\end{tabular}
\end{adjustbox}
\vspace{-1em}
\end{table}

\begin{figure*}
\vspace{-2mm}
\centering
\includegraphics[width=1\linewidth]{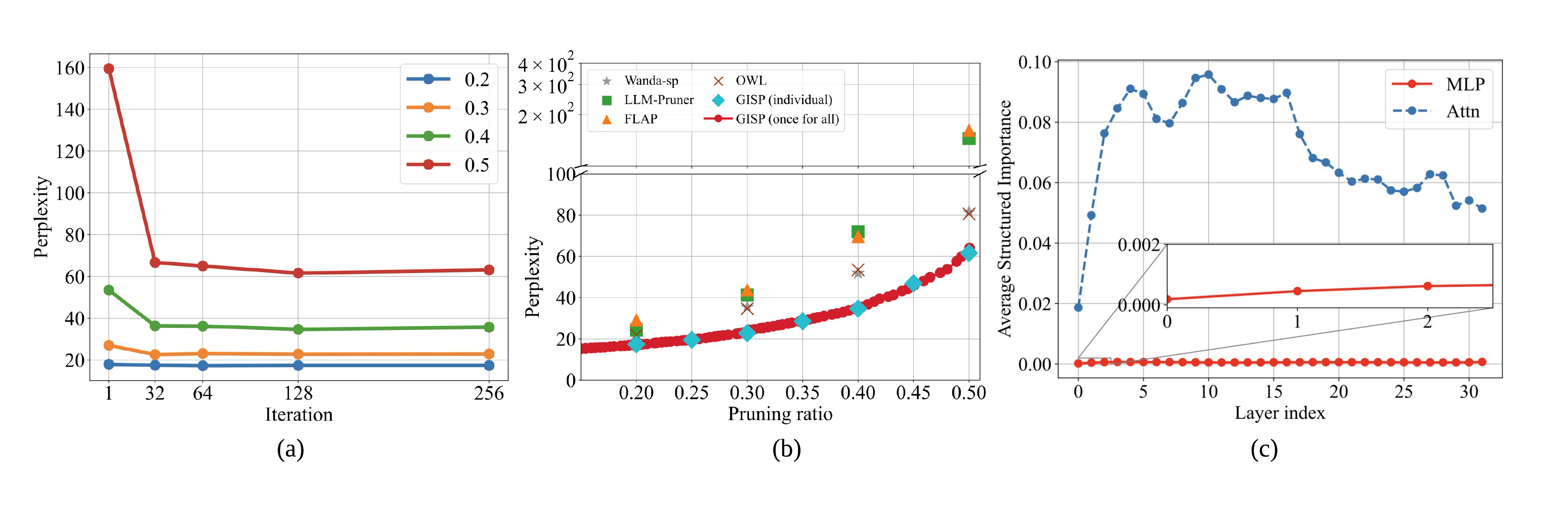}
\vspace{-8mm}
    \caption{(a) Perplexity analysis for various iteration settings. Iteration alleviates high-pruning-ratio perplexity collapse. (b) Perplexity analysis between GISP and other baselines. (c) Magnitude comparison between different types of structured weight importance.
    }\label{fig:ppl_its_analysis}
\vspace{-1.2em}
\end{figure*}

\begin{algorithm}[t]
\resizebox{\linewidth}{!}{%
\caption{Global Iterative Structured Pruning}
\label{alg:gisp} 
\begin{minipage}{1.2\linewidth}
\begin{algorithmic}[1]
\Require Network parameters $\theta_{0}$; global target pruning ratio $\rho$; iteration steps $n$;
calibration dataset $D=\{(x_i,y_i)\}_{i=1}^{N}$.
\Ensure Pruned model parameters $\theta_{n}$.
\State $\{\rho_t\}_{t=1}^{n} \gets \Call{RatioScheduler}{\rho, n}$
\For{$k \gets 1$ to $n$}
    \State $I(\theta_{k-1}) \gets |\frac{\partial \mathcal{D}}{\partial \theta_{k-1}} \odot \theta_{k-1}|$
    \Comment{First-order importance}
    \State $I(\theta_{k-1}) \gets \Call{SUMAggregate}{I(\theta_{k-1})}$
    \State $I({\theta_{k-1}}) \gets \frac{I({\theta_{\text{k-1}}}) }{\lvert \theta_{\text{k-1}}\rvert}$
    \Comment{Normalized Aggregation over heads / MLP channels}
    \State $\tau_k \gets \Call{TopK}{I(\theta_{k-1}), \rho_k}$
    \Comment{Global pruning threshold across different structures at ratio $\rho_k$}
    \State $m \gets \mathbf{1}\!\left[I(\theta_{k-1}) > \tau_k\right]$
    \Comment{Binary mask}
    \State $\theta_k \gets \theta_{k-1} \odot m$
    \Comment{Apply pruning mask}
\EndFor
\State \Return $\theta_n$
\end{algorithmic}
\end{minipage}
}
\vspace{-0.2em}
\end{algorithm}

\color{black}
\noindent\textbf{Global Iterative Structured Pruning.} Building upon the procedure detailed in Section \ref{sec:naive-one-shot-gp}, given a predefined number of iteration steps \( n \) and a target pruning ratio \( \rho \), GISP performs pruning iteratively using a small pruning ratio \( \rho_k \) at iteration \( k \), as shown in Fig~\ref{overview-fig}. The detailed algorithm block is provided in Algorithm \ref{alg:gisp}. To control pruning at each step, we use a linear scheduler that gradually increases the pruning ratio across iterations, ensuring that each iteration prunes the same number of structures. More details can be found in Sec.~\ref{sec:detailed_gisp}
\color{black}

\noindent\textbf{Empirical study.} For the iteration study, we vary the number of pruning steps (1, 32, 64, 128, and 256) across four target pruning ratios (20\%, 30\%, 40\%, and 50\%). For comparison with local pruning, we measure perplexity (PPL) and include four representative post‑training structured pruning baselines: two uniform local pruning methods (Wanda and LLM‑Pruner) and two non‑uniform local pruning methods (FLAP and OWL). All experiments are performed on the Llama2‑7B model using the C4 calibration dataset. The results are shown in Fig~\ref{fig:ppl_its_analysis}. We summarize our main findings below: 

\textit{1) \textbf{Iteration is key to global pruning in the high-sparsity regime}.} 
From Fig~\ref{fig:ppl_its_analysis}(a) and (b), we first observe that introducing iterative pruning alleviates the issue of global pruning at a high pruning ratio: even a coarse setting of 32 steps (equal to the layer count of Llama2-7B) is enough to cut the 50\%-pruning-ratio PPL by 92.82. As a result, the GISP can consistently achieve lower PPL compared to local pruning methods. 

\textit{2) \textbf{Global iterative pruning outperforms local baselines at scale.}} With iteration, global pruning consistently achieves lower perplexity than local pruning methods across all sparsity regimes (Fig.~\ref{fig:ppl_its_analysis}(b)). This establishes that iteration not only stabilizes global pruning but also makes it competitive against strong local baselines in the post-training LLM setting. Crucially, these gains are obtained without intermediate recovery or fine-tuning, demonstrating that iteration alone is effective.


\begin{table}[t]
\centering
\caption{Pruning time comparison across methods. ``Amortized time'' divides the total time by the number of usable subnetworks produced.}
\vspace{-2mm}
\label{tab:pruning_time}
\resizebox{1\linewidth}{!}{
\begin{tabular}{lccc}
\toprule
\textbf{Method} & \textbf{Usable sparsities} & \textbf{Total time (min)} & \textbf{Amortized time (min)} \\
\midrule
Wanda-sp & 4 & 7.10 & 1.78 \\
OWL & 4 & 13.90 & 3.48 \\
LLM-Pruner & 4 & 6.80 & 1.70 \\
GISP (ours) & 112 & 125.84 & 1.12 \\
\bottomrule
\end{tabular}}
\vspace{-5mm}
\end{table}

\subsubsection{Achieving ``Once-for-All'' and Amortizing the Iteration Cost}\label{sec:prune-once-for-all}
Iterative global pruning is computationally more demanding than local or one-shot pruning. Running such a computationally intensive procedure to obtain only a single subnetwork at a fixed sparsity level would be impractical in deployment. \Cref{tab:pruning_time} compares the wall-clock pruning time of several structured pruning methods under our experimental setup. 
While GISP requires a substantially longer total runtime due to its iterative steps, the \emph{amortized} cost per deployable subnetwork is comparable to, or even lower than, that of local methods once the once-for-all property is considered.

Moreover, in iterative global pruning, each step removes new self-attention heads and MLP channels based on the already-pruned model from the previous step, naturally forming a \textit{nested sub-network} structure~\cite{cai2019once}. This nested property and computational cost from iteration motivated us to wonder: 

\vspace{+0.3em}
\noindent\textit{Can GISP enable \textbf{once-for-all} pruning? In other words, if we iteratively prune toward a high target ratio (e.g., 50\%), can the intermediate sub-networks with lower pruning ratios (e.g., 20\%, 30\%) already perform well, thereby eliminating the need to conduct separate pruning runs for each individual pruning ratio? }

\vspace{+0.3em}

To investigate this, we conduct a single iterative pruning procedure on Llama2-7B, targeting 50\% sparsity over 112 iterations. We saved the intermediate pruned model at every step and evaluated its PPL. The results are presented in Fig~\ref{fig:ppl_its_analysis}(b). The relationship between perplexity and the pruning ratio is remarkably smooth, indicating a stable and well-behaved pruning trajectory. Crucially, the performance of the intermediate models at different pruning ratios is on par with the performance of models generated from individual, shorter pruning runs (marked as "individual" variant) tuned specifically for those respective targets. To this end, this demonstrates the \textit{once-for-all} capacity of GISP.  

It is important to note that this "once-for-all" capability is a unique advantage of the iterative global pruning. It enables practitioners to obtain an entire Pareto frontier of accuracy-vs-sparsity models from a single computational investment, offering immense practical flexibility. This property is not achievable with local pruning methods. As formulated in Eq. 1, local pruning is a layer-wise optimization that requires the target pruning ratio for each layer to be specified in advance. Consequently, creating models for different sparsity levels necessitates entirely separate and independent pruning runs.

\subsection{GISP as a Task-Specific Pruner}

As discussed in Sec.~\ref{sec:preliminary}, local pruning remains task-agnostic because its layer-wise reconstruction objective cannot align with downstream goals, even when calibration data carries task-specific information. In contrast, global pruning defines importance with respect to a model-level loss, offering the potential for task alignment. We now instantiate and validate this property in GISP.

\noindent\textbf{Objective-level formulation.}
Because GISP evaluates importance with respect to a \emph{model-level} loss (Eq.~\ref{eq:score_glocal}), 
we can instantiate a \emph{task-aligned} objective by replacing the loss in Eq.~\ref{eqt:global_pruning} with a task-specific target $L_{\text{task}}$. 
Our importance reduces to the same first-order form with a different objective:
\begin{equation}
\small
I_{W} \;=\; \bigl|\langle \nabla_{W} L_{\text{task}},\, W \rangle \bigr| \, .
\label{eq:task_aligned_importance}
\end{equation}
This simple substitution turns GISP into a \emph{task-specific} pruner while remaining post-training and structure-aware.

\noindent\textbf{Two instantiations.}
We consider two common families of $L_{\text{task}}$ that match our evaluation tasks:

(i) \textbf{Perplexity loss} for text generation (language modeling), where $L_{\text{task}}{=}$token-level cross-entropy on an open-domain (e.g., C4) or in-domain (e.g., GSM8K) corpus; To be specific, the importance metrics are obtained from objective:
\begin{equation}
\label{eqt:ppl_loss_def}
    \small
L = -\frac{1}{N}\sum_{i=1}^{N}\log p(x_i|x_{<i})
\end{equation}
where $L$ is the loss function, $N$ is the number of tokens and $p(x_i|x_{<i})$ is the probability of token $x_i$ given all previous tokens.

(ii) \textbf{Margin loss} for decision-oriented, multi-option QA. For example, the CMQA dataset differs from pure language modeling in that each question is paired with one correct (positive) and multiple incorrect (negative) answers. During inference, the model ranks each ‘Question + Answer’ pair by perplexity and selects the answer with the lowest score. Simply minimizing perplexity on positive answers is insufficient, as pruning may disproportionately reduce the loss of negative candidates relative to the correct one, causing the model to choose an incorrect answer even if the correct answer’s loss remains largely unchanged. In other words, \textit{the actual factor of classification performance is the model's ability to distinguish correct from incorrect answers (the decision boundary).}

To preserve the model’s decision boundary, we define a margin-based importance using a task-formatted calibration set:
\begin{equation}
\label{eqt:margin_loss_def}
    \small
  I_{W_i^j}
= \left\lvert
\Bigl(\frac{\partial L_{+}}{\partial W_i^j}
\;-\;
\frac{\partial L_{-}}{\partial W_i^j}\Bigr)
\,W_i^j
\right\rvert  
\end{equation}

Where $L_{+}$ denotes the average loss on positive labels and $L_{-}$ denotes the average loss on negative labels. 
Intuitively, Eq.~\eqref{eqt:margin_loss_def} preserves the loss gap between correct and incorrect candidates, aligning pruning with task decisions. We will examine the effectiveness of GISP as a task-specific pruner in Sec.~\ref{sec:task-specific_ex}.
Importantly, such a transition from a perplexity-based loss to a task-specific loss is not feasible for local pruning methods, which rely on layer-wise MSE loss for importance estimation.

\begin{table}[t]
\centering
\caption{Comparison of different pruning methods on perplexity and CMQA downstream reasoning accuracy.}
\vspace{-2mm}
\label{tab:experiments_llama2}
\centering
\resizebox{1\linewidth}{!}{
\begin{tabular}{lcccccc}
\toprule
\multirow{3}{*}{\parbox{1.5cm}{\textbf{Pruning\\Ratio}}} 
& \multirow{3}{*}{\textbf{Method}} 
& \multicolumn{2}{c}{\textbf{Perplexity on Wikitext2} $\downarrow$} 
& \multicolumn{2}{c}{\textbf{Downstream ACC (\%)} $\uparrow$} \\
\cmidrule(lr){3-4} \cmidrule(lr){5-6}
& & \multicolumn{2}{c}{Llama2} & \multicolumn{2}{c}{Llama2} \\
\cmidrule(lr){3-4} \cmidrule(lr){5-6}
& & 7B & 13B & 7B & 13B \\
\midrule
0\% & Dense & 12.19 & 10.98 & 66.68 & 69.19 \\
\midrule
\multirow{5}{*}{20\%} 
& Wanda-sp & 22.71 & \textbf{14.64} & 61.77 & 66.30 \\
& LLM-Pruner & 24.25 & 19.99 & 50.95 & 57.30 \\
& FLAP & 29.19 & 16.95 & 61.27 & 65.00 \\
& ShortGPT & 43.88 & 19.95 & 55.75 & 60.84 \\
& OWL & 21.80 & 14.76 & 62.64 & 66.98 \\
& GISP (ours) & \textbf{17.01} & 15.10 & \textbf{63.46} & \textbf{67.61} \\
\midrule
\multirow{5}{*}{30\%} 
& Wanda-sp & 35.43 & 19.73 & 57.14 & 62.69 \\
& LLM-Pruner & 41.24 & 28.47 & 41.37 & 46.26 \\
& FLAP & 43.75 & 21.32 & 56.90 & 63.28 \\
& ShortGPT & 126.42 & 84.84 & 50.01 & 56.86 \\
& OWL & 34.64 & \textbf{19.02} & 58.33 & 63.27 \\
& GISP (ours) & \textbf{24.27} & 19.53 & \textbf{60.68} & \textbf{66.12} \\
\midrule
\multirow{5}{*}{40\%} 
& Wanda-sp & 51.85 & 32.91 & 50.12 & 59.11 \\
& LLM-Pruner & 71.93 & 50.01 & 39.13 & 41.99 \\
& FLAP & 69.64 & 37.76 & 53.01 & 59.89 \\
& ShortGPT & 189.17 & 92.38 & 45.35 & 48.73 \\
& OWL & 53.47 & 31.13 & 51.50 & 60.84 \\
& GISP (ours) & \textbf{34.54} & \textbf{26.56} & \textbf{55.28} & \textbf{63.34} \\
\midrule
\multirow{5}{*}{50\%} 
& Wanda-sp & 81.47 & 64.17 & 43.52 & 51.60 \\
& LLM-Pruner & 144.99 & 86.34 & 38.62 & 40.92 \\
& FLAP & 161.84 & 66.38 & 47.84 & 56.29 \\
& ShortGPT & 387.94 & 276.08 & 41.75 & 41.75 \\
& OWL & 80.59 & 65.28 & 44.82 & 54.17 \\
& GISP (ours) & \textbf{64.07} & \textbf{42.07} & \textbf{48.54} & \textbf{57.50} \\
\bottomrule
\end{tabular}}
\vspace{-1em}
\end{table}

\begin{table}[t]
\centering
\caption{Comparison of different pruning methods for advanced models.}
\label{tab:exp_advanced_models}
\vspace{-2mm}
\centering
\resizebox{1\linewidth}{!}{
\begin{tabular}{lcccccc}
\toprule
\multirow{3}{*}{\parbox{1.5cm}{\textbf{Pruning\\Ratio}}} 
& \multirow{3}{*}{\textbf{Method}} 
& \multicolumn{2}{c}{\textbf{Perplexity on Wikitext2} $\downarrow$} 
& \multicolumn{2}{c}{\textbf{Downstream ACC (\%)} $\uparrow$} \\
\cmidrule(lr){3-4} \cmidrule(lr){5-6}
& & Llama3 & Mistral-0.3 & Llama3 & Mistral-0.3 \\
& & 8B & 7B & 8B & 7B \\
\midrule
0\% & Dense & 14.14 & 15.14 & 69.99 & 70.47 \\
\midrule
\multirow{5}{*}{20\%} 
& Wanda-sp & 29.92 & 20.42 & 57.45 & 64.39 \\
& LLM-Pruner & \textbf{23.21} & \textbackslash & 56.51 & \textbackslash \\
& ShortGPT & 118.62 & 52.74 & 57.68 & 57.75 \\
& OWL & 29.49 & 19.98 & 59.95 & 65.87 \\
& GISP (ours) & 24.18 & \textbf{18.17} & \textbf{65.28} & \textbf{66.60} \\
\midrule
\multirow{5}{*}{30\%} 
& Wanda-sp & 48.83 & 32.61 & 52.03 & 58.24 \\
& LLM-Pruner & 37.78 & \textbackslash & 47.46 & \textbackslash \\
& ShortGPT & 3972.28 & 599.82 & 43.53 & 41.10 \\
& OWL & 47.90 & 31.82 & 52.24 & 58.54 \\
& GISP (ours) & \textbf{31.73} & \textbf{25.58} & \textbf{59.66} & \textbf{63.48} \\
\midrule
\multirow{5}{*}{40\%} 
& Wanda-sp & 81.67 & 55.41 & 43.61 & 51.89 \\
& LLM-Pruner & 67.58 & \textbackslash & 41.82 & \textbackslash \\
& ShortGPT & 1576.47 & 909.21 & 43.37 & 39.68 \\
& OWL & 87.01 & 47.85 & 44.87 & 54.36 \\
& GISP (ours) & \textbf{46.10} & \textbf{34.31} & \textbf{53.51} & \textbf{58.30} \\
\midrule
\multirow{5}{*}{50\%} 
& Wanda-sp & 133.29 & 79.41 & 41.32 & 44.38 \\
& LLM-Pruner & 125.91 & \textbackslash & 39.67 & \textbackslash \\
& ShortGPT & 4135.73 & 1091.73 & 41.19 & 38.73 \\
& OWL & 130.77 & 76.20 & 41.86 & 46.09 \\
& GISP (ours) & \textbf{79.42} & \textbf{58.16} & \textbf{45.68} & \textbf{49.79} \\
\bottomrule
\end{tabular}}
\vspace{-1.5em}
\end{table}

\section{Experiments}\label{sec:exp}
\textbf{Models and Evaluation.} We evaluate GISP on the popular Llama2-7B/13B~\cite{touvron2023llama2openfoundation}, Llama3-8B~\cite{grattafiori2024llama3herdmodels}, Mistral-0.3-7B~\cite{jiang2023mistral}, and two reasoning model DeepSeek-R1-Distill-Llama-3-8B~\cite{deepseekai2025deepseekr1incentivizingreasoningcapability} and Qwen3-8B~\cite{yang2025qwen3technicalreport}. Following previous work \cite{ma2023llm,an2023fluctuationbasedadaptivestructuredpruning}, we evaluate the pruned model on three categories of tasks: the perplexity metric on Wikitext2 \cite{merity2016pointersentinelmixturemodels} text generation, exact-match accuracy on math benchmark GSM8K \cite{cobbe2021trainingverifierssolvemath} that require step-by-step reasoning, and post-training accuracy on commonsense reasoning (CMQA), which includes  BoolQ \cite{clark-etal-2019-boolq}, PIQA \cite{Bisk2020piqa}, HellaSwag \cite{zellers2019hellaswag}, WinoGrande \cite{ai2:winogrande}, ARC-Easy \cite{allenai:arc}, ARC-Challenge \cite{allenai:arc}, and OpenbookQA \cite{OpenBookQA2018}. 
We report average CMQA accuracy in this section, and detailed task-wise accuracy in Sec.~\ref{sec:detailed_downstream}. \textcolor{black}{Additionally, we further evaluate the proposed GISP on the domain-specific MedQA benchmark \cite{jin2020diseasedoespatienthave}, including medical exam-style multiple-choice questions. The result can be found in Sec.~\ref{sec:app_medqa}.} \textcolor{black}
{For calibration, we use different datasets for perplexity evaluation and downstream accuracy: 1) for perplexity evaluation on WikiText-2, we follow the standard setup and use C4 for calibration, 2) for downstream accuracy, we instead use task-specific calibration data. Specifically, in CMQA, we adopt a unified multi-task pruning setting by constructing a single calibration set from the training splits of all seven tasks under a shared 512K-token budget, evenly allocated across tasks. One shared model is pruned using this combined calibration set and then evaluated on each task. For each sampled question, we retain the task-formatted answer candidates. Under the margin-based objective, the correct answer is treated as positive, and the remaining candidates are treated as negatives. For GSM8K, we consider both C4 and the GSM8K training set for calibration. In particular, we sample 500 sequences from C4, each of length 2048 tokens, and 500 randomly sampled question-solution pairs from the GSM8K training set.}


\begin{table*}[ht]
  \centering
  \caption{CMQA Accuracy of GISP on all seven tasks under different calibration datasets and pruning ratios. The best results are marked in \textbf{bold}.}
    \vspace{-0.5em}
  \label{tab:calibration_full_gisp}
    \resizebox{0.85\linewidth}{!}{
  \begin{tabular}{lccccccccc}
    \toprule
\textbf{Calibration Dataset}   & \textbf{Pruning Ratio} & \textbf{BoolQ} & \textbf{PIQA}  & \textbf{Hellaswag} & \textbf{WinoGrande} & \textbf{ARC-E} & \textbf{ARC-C} & \textbf{OBQA}  & \textbf{AVG}   \\
    \midrule
                \multirow{4}{*}{C4 + Perplexity}      &      20\%     & 73.30 & 78.45 & 73.09     & 68.11      & 67.59  & 42.92  & 42.00 & 63.64 \\
                 &                              30\%     & 69.20 & 76.71 & 69.68     & 65.98      & 62.67  & 37.46  & 40.80 & 60.36 \\
                 &                              40\%     & 65.14 & 73.67 & 62.80     & 61.64      & 54.55  & 33.87  & 36.80 & 55.49 \\
                 &                              50\%     & 58.41 & 68.28 & 50.79     & 58.64      & 44.02  & 28.41  & 32.20 & 48.68 \\
    \cmidrule(l){1-10}
                \multirow{4}{*}{CMQA + Perplexity}         & 20\%     & \textbf{80.80} & 77.86 & 76.39     & 71.82      & 74.75  & \textbf{46.33}  & 42.20 & 67.16 \\
                 &                               30\%     & 80.83 & 75.52 & 71.91     & 69.69      & 72.22  & 44.71  & 41.60 & 65.21 \\
                 &                               40\%     & 79.54 & 72.52 & 63.36     & 67.88      & 67.85  & 41.81  & 39.20 & 61.74 \\
                 &                               50\%     & \textbf{76.18} & 67.52 & 51.67     & 61.56      & 59.47  & 36.77  & 36.40 & 55.65 \\
    \cmidrule(l){1-10}
                  \multirow{4}{*}{CMQA + Margin} & 20\% & 80.28 & \textbf{79.00} & \textbf{76.83} & \textbf{72.22}      & \textbf{75.17}  & 45.99  & \textbf{43.80} & \textbf{67.61} \\
                 &                               30\%     & \textbf{81.16} & \textbf{76.77} & \textbf{72.87}     & \textbf{71.59}      & \textbf{72.94}  & \textbf{45.73}  & \textbf{41.80} & \textbf{66.12} \\
                 &                               40\%     & \textbf{80.00} & \textbf{73.83} & \textbf{65.79}     & \textbf{70.09}      & \textbf{70.16}  & \textbf{43.52}  & \textbf{40.00} & \textbf{63.34} \\
                 &                               50\%     & 72.97 & \textbf{69.91} & \textbf{55.15}     & \textbf{65.59}      & \textbf{63.09}  & \textbf{38.23}  & \textbf{37.60} & \textbf{57.50} \\
    \bottomrule
  \end{tabular}
  }
  \vspace{-1em}
\end{table*}

\noindent\textbf{Baselines.} We compare GISP with four local pruning approaches in two main categories: (1) local uniform baselines, including Wanda-sp \cite{sun2024simpleeffectivepruningapproach,an2023fluctuationbasedadaptivestructuredpruning}, LLM-Pruner \cite{ma2023llm}; and (2) local non-uniform baselines: FLAP~\cite{an2023fluctuationbasedadaptivestructuredpruning}, OWL~\cite{yin2025outlierweighedlayerwisesparsity} on Wanda-sp. Additionally, we compare against a layer-wise pruning approach, ShortGPT~\cite{men2024shortgptlayerslargelanguage}. \textcolor{black}{We also include SparseLLM~\cite{bai2024sparsellmglobalpruningpretrained}, a global pruning method originally proposed for unstructured pruning, by implementing a structured variant for comparison under our setting; the detailed results are provided in Appendix~\ref{sec:sparsellm_comparison}.}
The iteration step in GISP is set to 112 in models with a 7-8B scale, and 280 in 13B to maintain the close iteration stride with these smaller variants. The detailed experimental setup is described in the Sec.~\ref{sec:detailed_exp_setup}.

\subsection{Experimental Results}\label{sec:exp_main_result} 
\textbf{Perplexity of text generation tasks.} Table~\ref{tab:experiments_llama2} and Table~\ref{tab:exp_advanced_models}  present the experimental results on the perplexity (PPL) of WikiText2 across four target pruning ratio levels. First of all, compared to the five baselines on dense Llama2-7B, 13B models, GISP  achieves a clear lower PPL in most cases. Specifically, the improvement is particularly significant at the higher sparsity level (e.g., 40\%, and 50\%). 
Moreover, for multi‑query attention–based LLMs such as Llama3‑8B and Mistral‑0.3, we observe the same consistent trend~\footnote{We exclude the results of FLAP on Llama3-8B and Mistral-0.3, and leave LLM-Pruner on Mistral-0.3 as blank since it requires non-trivial, architecture-specific modifications, and these models are not officially supported in their open-sourced code.}. 

\noindent\textbf{Downstream accuracy of commonsense reasoning tasks.} Table~\ref{tab:experiments_llama2} and Table~\ref{tab:exp_advanced_models} summarize the accuracy results of CMQA under downstream task evaluations. Note that Downstream Accuracy is evaluated by using CMQA calibration. We observe that on downstream tasks, GISP consistently achieves higher accuracy across all models and pruning ratios, with particularly strong gains at higher pruning levels, indicating its strength as a task-specific pruner. 

\noindent\textbf{Exact-match accuracy of answer-generation tasks.}
While CMQA evaluates multiple-choice reasoning, we further validate GISP on the arithmetic reasoning benchmark GSM8K, which follows a text-generation format and is evaluated by the presence of the gold answer in the generated output (Gold ACC). Table~\ref{tab:gsm8k_8shot_final} compares different pruning methods and calibration datasets under 8-shot evaluation on DeepSeek-R1-Distill-Llama-3-8B and Qwen3-8B. The same trend holds across both models: task-informed calibration substantially improves GISP over generic C4 calibration, and GISP consistently achieves stronger Gold ACC than Wanda-sp at the same sparsity levels. These results further support the effectiveness of task-aligned calibration for generative reasoning tasks. Notably, pruning reasoning-oriented LLMs remains challenging for current methods \citep{zhang2025reasoningmeetscompressionunderstanding,sui2025stopoverthinkingsurveyefficient}; under default C4 calibration, Wanda-sp even collapses to 0\% accuracy at 20\% pruning on DeepSeek-R1-Distill-Llama-3-8B.

\begin{table}[t]
\centering
\caption{Comparison of different pruning methods on GSM8K (8-shot) Gold ACC (\%).}
  \vspace{-0.5em}
\label{tab:gsm8k_8shot_final}
\resizebox{0.92\linewidth}{!}{
\begin{tabular}{llccc}
\toprule
\textbf{Model} & \textbf{Calibration} & \textbf{Method} & \textbf{Ratio} & \textbf{Gold ACC (\%)} \\
\midrule

\multirow{9}{*}{\parbox{2.2cm}{\centering DeepSeek-R1-\\Distill-Llama-3-8B}}
& N/A & Dense & 0\% & 73.54\% \\
\cmidrule(lr){2-5}

& \multirow{4}{*}{C4}
& Wanda-sp & 20\% & 0.00\% \\
\cmidrule(lr){3-5}
& & \multirow{3}{*}{GISP} & 20\% & 25.25\% \\
& &  & 30\% & 14.33\% \\
& &  & 40\% & 5.46\% \\
\cmidrule(lr){2-5}

& \multirow{4}{*}{GSM8K}
& Wanda-sp & 20\% & 29.19\% \\
\cmidrule(lr){3-5}
& & \multirow{3}{*}{GISP} & 20\% & 67.93\% \\
& &  & 30\% & 50.80\% \\
& &  & 40\% & 31.84\% \\
\midrule

\multirow{9}{*}{\parbox{2.2cm}{\centering Qwen3-8B}}
& N/A & Dense & 0\% & 96.74\% \\
\cmidrule(lr){2-5}

& \multirow{4}{*}{C4}
& \multirow{2}{*}{Wanda-sp} & 20\% & 25.40\% \\
& &  & 30\% & 2.81\% \\
\cmidrule(lr){3-5}
& & \multirow{2}{*}{GISP} & 20\% & 47.31\% \\
& &  & 30\% & 9.25\% \\
\cmidrule(lr){2-5}

& \multirow{4}{*}{GSM8K}
& \multirow{2}{*}{Wanda-sp} & 20\% & 85.90\% \\
& &  & 30\% & 16.38\% \\
\cmidrule(lr){3-5}
& & \multirow{2}{*}{GISP} & 20\% & 90.52\% \\
& &  & 30\% & 79.68\% \\
\bottomrule
\end{tabular}
}
\vspace{-1.2em}
\end{table}

\subsection{Task-specific Property of GISP}\label{sec:task-specific_ex}
\noindent\textbf{Ablation across seven CMQA tasks.} We use CMQA to examine the task-specific property of GISP in a unified multi-task pruning setting described at the beginning of Sec.~\ref {sec:exp}. We report detailed results on all tasks across pruning ratios of 20\% to 50\%. 
Table~\ref{tab:calibration_full_gisp} summarizes two consistent trends: 
(1) \textbf{Task-informed calibration helps even with a perplexity target}: replacing C4 with CMQA data under the same perplexity objective yields gains at all ratios, indicating that GISP is an intrinsic task-specific pruner that can actively benefit from task signals from the calibration dataset.
(2) \textbf{Task-specific loss target brings further improvements}: switching from perplexity to the proposed margin objective (Eq.~\ref{eqt:margin_loss_def}) provides additional, consistent accuracy gains, especially at higher pruning ratios. \textcolor{black}{These trends hold across tasks, showing that GISP can capture cross-task sensitivity within a shared structured subnetwork and supporting GISP as a practical task-specific pruner.}

\noindent\textbf{Post-pruning fine-tuning performance.} We also examine whether the task-aligned subnetworks identified by GISP remain favorable for further downstream adaptation. Starting from the 20\% and 40\% pruned Llama2-7B checkpoints, we fine-tune the pruned models with LoRA on an additional 2M CMQA samples under identical configurations. Table~\ref{tab:lora_after_pruning} shows that GISP consistently achieves higher final accuracy after LoRA adaptation at both sparsity levels. This suggests that GISP not only yields stronger pruned subnetworks but also provides a better starting point for subsequent task adaptation.

\begin{table}[t]
\centering
\small
\caption{Average CMQA accuracy after LoRA fine-tuning on an additional 2M CMQA samples.}
\vspace{-2mm}
\label{tab:lora_after_pruning}
\begin{tabular}{lcccc}
\toprule
& \multicolumn{2}{c}{20\% sparsity} & \multicolumn{2}{c}{40\% sparsity} \\
\cmidrule(lr){2-3}\cmidrule(lr){4-5}
Method & Before & After & Before & After \\
\midrule
Wanda-sp    & 61.84 & 67.43 & 50.09 & 58.66 \\
GISP (ours) & \textbf{63.46} & \textbf{68.93} & \textbf{55.28} & \textbf{60.52} \\
\bottomrule
\end{tabular}
\vspace{-2mm}
\end{table}

\begin{table}[t]
\centering
\small
\caption{CMQA ACC Comparison between limited (1\%) and default (100\%) budgets on Llama2-7B.}
\vspace{-2mm}
\resizebox{0.9\linewidth}{!}{
\label{tab:calib_budget_anchor}
\begin{tabular}{lcccc}
\toprule
Token budget & 20\% & 30\% & 40\% & 50\% \\
\midrule
5,120 (1\%)      & 63.86 & 60.46 & 55.83 & 49.20 \\
512,000 (100\%)  & 63.46 & 60.68 & 55.28 & 48.54 \\
\midrule
Absolute diff.       & 0.40  & 0.22  & 0.55  & 0.66  \\
\bottomrule
\end{tabular}
}
\vspace{-4mm}
\end{table}

\begin{figure}[t]
\centering
\includegraphics[width=1\linewidth]{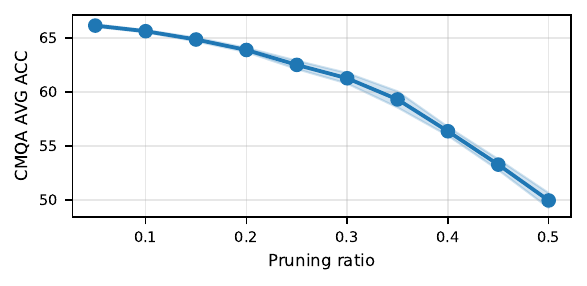}
\vspace{-9mm}
    \caption{Robustness of the iterative pruning trajectory from GISP under reduced calibration budgets. The shaded region indicates standard deviation.}
\vspace{-5mm}
\label{fig:calib_robust}
\end{figure}

\color{black}
\subsection{Calibration Robustness of GISP}
\label{sec:calib_robust}

In this section, we study calibration robustness from two perspectives: robustness to calibration budget and robustness to calibration sampling. To be specific, we conduct GISP on Llama2-7B with the same settings in Sec.~\ref{sec:exp_main_result} and evaluate on CMQA accuracy.

\noindent\textbf{Robustness to calibration budget.}
We first analyze how varying the calibration budget affects the full pruning trajectory. Starting from the default 512K-token CMQA calibration budget, we subsample the same calibration pool to 1\%, 20\%, 40\%, 60\%, and 80\%, keep all other settings fixed, rerun the full iterative pruning procedure under each reduced-budget setting, and evaluate 10 sparsity points from 5\% to 50\%. Figure~\ref{fig:calib_robust} summarizes the mean and standard deviation of the resulting pruning trajectories across these reduced-budget settings, while the detailed results can be found in Sec.~\ref{sec:app_detailed_robust}. As shown in the figure, the overall pruning trajectory remains smooth throughout the full range, and the variation stays small, with the maximum standard deviation below 0.75. This suggests that GISP remains robust even when the available calibration budget is substantially reduced.

Interestingly, during this analysis, we also observe that even at 1\% of the default budget, the resulting CMQA accuracy already remains close to the full-budget setting at the commonly used 20\%, 30\%, 40\%, and 50\% pruning ratios. Table~\ref{tab:calib_budget_anchor} provides a direct comparison between the 1\% and 100\% budgets. The absolute difference is at most 0.66 points, which strongly supports the robustness of GISP under limited calibration data.

\noindent\textbf{Robustness to calibration sampling.}
To isolate the effect of sample composition from budget size, we additionally resample the calibration subset with random seeds of 0-2, obtaining $63.44{\pm}0.07$, $60.50{\pm}0.25$, $55.67{\pm}0.36$, and $49.11{\pm}0.56$ average accuracy at 20\%, 30\%, 40\%, and 50\% pruning, respectively. These results indicate that GISP is also stable under different calibration samples, with a standard deviation of at most 0.56 points.
\color{black}

\section{Conclusions}
In this work, we propose GISP, a simple yet effective global iterative structured pruning method for LLMs. GISP prunes globally and iteratively, enabling more flexible, task-aware pruning. It supports once-for-all pruning across multiple sparsity levels and naturally incorporates loss functions tailored to downstream tasks to guide weight importance. Experiments conducted on multiple model families demonstrate clear performance gains compared to prior works, excelling as a task-specific pruner, particularly at high sparsity.

\section*{Limitations}
One limitation of our method is that, due to its reliance on gradient-based weight importance estimation, it can incur relatively high memory and computational costs. To address this, one could integrate parameter-efficient fine-tuning (PEFT) techniques to accelerate importance computations and reduce the memory footprint—a direction we leave for future work. Additionally, while GISP is designed to be architecture-agnostic and shows promising results on multi-query attention (MQA)-based architectures, we have not yet evaluated it on Mixture-of-Experts (MoE) models due to their significantly larger scale. Extending GISP to MoE architectures remains a valuable direction for future exploration. 

\bibliography{main}

@article{ma2023llm,
  title={Llm-pruner: On the structural pruning of large language models},
  author={Ma, Xinyin and Fang, Gongfan and Wang, Xinchao},
  journal={Advances in neural information processing systems},
  volume={36},
  pages={21702--21720},
  year={2023}
}

@article{touvron2023llama,
  title={Llama: Open and efficient foundation language models},
  author={Touvron, Hugo and Lavril, Thibaut and Izacard, Gautier and Martinet, Xavier and Lachaux, Marie-Anne and Lacroix, Timoth{\'e}e and Rozi{\`e}re, Baptiste and Goyal, Naman and Hambro, Eric and Azhar, Faisal and others},
  journal={arXiv preprint arXiv:2302.13971},
  year={2023}
}

@article{molchanov2016pruning,
  title={Pruning convolutional neural networks for resource efficient inference},
  author={Molchanov, Pavlo and Tyree, Stephen and Karras, Tero and Aila, Timo and Kautz, Jan},
  journal={arXiv preprint arXiv:1611.06440},
  year={2016}
}

@misc{touvron2023llama2openfoundation,
      title={Llama 2: Open Foundation and Fine-Tuned Chat Models}, 
      author={Hugo Touvron and others},
      year={2023},
      eprint={2307.09288},
      archivePrefix={arXiv},
      primaryClass={cs.CL},
      url={https://arxiv.org/abs/2307.09288}, 
}

@misc{an2023fluctuationbasedadaptivestructuredpruning,
      title={Fluctuation-based Adaptive Structured Pruning for Large Language Models}, 
      author={Yongqi An and Xu Zhao and Tao Yu and Ming Tang and Jinqiao Wang},
      year={2023},
      eprint={2312.11983},
      archivePrefix={arXiv},
      primaryClass={cs.CL},
      url={https://arxiv.org/abs/2312.11983}, 
}

@misc{sun2024simpleeffectivepruningapproach,
      title={A Simple and Effective Pruning Approach for Large Language Models}, 
      author={Mingjie Sun and Zhuang Liu and Anna Bair and J. Zico Kolter},
      year={2024},
      eprint={2306.11695},
      archivePrefix={arXiv},
      primaryClass={cs.CL},
      url={https://arxiv.org/abs/2306.11695}, 
}

@inproceedings{clark-etal-2019-boolq,
    title = "{B}ool{Q}: Exploring the Surprising Difficulty of Natural Yes/No Questions",
    author = "Clark, Christopher  and
      Lee, Kenton  and
      Chang, Ming-Wei  and
      Kwiatkowski, Tom  and
      Collins, Michael  and
      Toutanova, Kristina",
    booktitle = "Proceedings of the 2019 Conference of the North {A}merican Chapter of the Association for Computational Linguistics: Human Language Technologies, Volume 1 (Long and Short Papers)",
    month = jun,
    year = "2019",
    address = "Minneapolis, Minnesota",
    publisher = "Association for Computational Linguistics",
    url = "https://aclanthology.org/N19-1300",
    doi = "10.18653/v1/N19-1300",
    pages = "2924--2936",
}

@inproceedings{Bisk2020piqa,
  author = {Yonatan Bisk and Rowan Zellers and
            Ronan Le Bras and Jianfeng Gao
            and Yejin Choi},
  title = {PIQA: Reasoning about Physical Commonsense in
           Natural Language},
  booktitle = {Thirty-Fourth AAAI Conference on
               Artificial Intelligence},
  year = {2020},
}

@inproceedings{zellers2019hellaswag,
    title={HellaSwag: Can a Machine Really Finish Your Sentence?},
    author={Zellers, Rowan and Holtzman, Ari and Bisk, Yonatan and Farhadi, Ali and Choi, Yejin},
    booktitle ={Proceedings of the 57th Annual Meeting of the Association for Computational Linguistics},
    year={2019}
}

@misc{ai2:winogrande,
      title={WinoGrande: An Adversarial Winograd Schema Challenge at Scale}, 
      author={Keisuke Sakaguchi and Ronan Le Bras and Chandra Bhagavatula and Yejin Choi},
      year={2019},
      eprint={1907.10641},
      archivePrefix={arXiv},
      primaryClass={cs.CL}
}

@article{allenai:arc,
      author    = {Peter Clark  and Isaac Cowhey and Oren Etzioni and Tushar Khot and
                    Ashish Sabharwal and Carissa Schoenick and Oyvind Tafjord},
      title     = {Think you have Solved Question Answering? Try ARC, the AI2 Reasoning Challenge},
      journal   = {arXiv:1803.05457v1},
      year      = {2018},
}

@inproceedings{OpenBookQA2018,
 title={Can a Suit of Armor Conduct Electricity? A New Dataset for Open Book Question Answering},
 author={Todor Mihaylov and Peter Clark and Tushar Khot and Ashish Sabharwal},
 booktitle={EMNLP},
 year={2018}
}

@misc{merity2016pointersentinelmixturemodels,
      title={Pointer Sentinel Mixture Models}, 
      author={Stephen Merity and Caiming Xiong and James Bradbury and Richard Socher},
      year={2016},
      eprint={1609.07843},
      archivePrefix={arXiv},
      primaryClass={cs.CL},
      url={https://arxiv.org/abs/1609.07843}, 
}

@article{cai2019once,
  title={Once-for-all: Train one network and specialize it for efficient deployment},
  author={Cai, Han and Gan, Chuang and Wang, Tianzhe and Zhang, Zhekai and Han, Song},
  journal={arXiv preprint arXiv:1908.09791},
  year={2019}
}

@misc{frantar2023sparsegptmassivelanguagemodels,
      title={SparseGPT: Massive Language Models Can Be Accurately Pruned in One-Shot}, 
      author={Elias Frantar and Dan Alistarh},
      year={2023},
      eprint={2301.00774},
      archivePrefix={arXiv},
      primaryClass={cs.LG},
      url={https://arxiv.org/abs/2301.00774}, 
}

@article{blalock2020state,
  title={What is the state of neural network pruning?},
  author={Blalock, Davis and Gonzalez Ortiz, Jose Javier and Frankle, Jonathan and Guttag, John},
  journal={Proceedings of machine learning and systems},
  volume={2},
  pages={129--146},
  year={2020}
}

@misc{men2024shortgptlayerslargelanguage,
      title={ShortGPT: Layers in Large Language Models are More Redundant Than You Expect}, 
      author={Xin Men and Mingyu Xu and Qingyu Zhang and Bingning Wang and Hongyu Lin and Yaojie Lu and Xianpei Han and Weipeng Chen},
      year={2024},
      eprint={2403.03853},
      archivePrefix={arXiv},
      primaryClass={cs.CL},
      url={https://arxiv.org/abs/2403.03853}, 
}

@misc{bai2024sparsellmglobalpruningpretrained,
      title={SparseLLM: Towards Global Pruning for Pre-trained Language Models}, 
      author={Guangji Bai and Yijiang Li and Chen Ling and Kibaek Kim and Liang Zhao},
      year={2024},
      eprint={2402.17946},
      archivePrefix={arXiv},
      primaryClass={cs.CL},
      url={https://arxiv.org/abs/2402.17946}, 
}

@misc{yang2023globalvisiontransformerpruning,
      title={Global Vision Transformer Pruning with Hessian-Aware Saliency}, 
      author={Huanrui Yang and Hongxu Yin and Maying Shen and Pavlo Molchanov and Hai Li and Jan Kautz},
      year={2023},
      eprint={2110.04869},
      archivePrefix={arXiv},
      primaryClass={cs.CV},
      url={https://arxiv.org/abs/2110.04869}, 
}

@misc{diao2023pruningdeepneuralnetworks,
      title={Pruning Deep Neural Networks from a Sparsity Perspective}, 
      author={Enmao Diao and Ganghua Wang and Jiawei Zhan and Yuhong Yang and Jie Ding and Vahid Tarokh},
      year={2023},
      eprint={2302.05601},
      archivePrefix={arXiv},
      primaryClass={cs.LG},
      url={https://arxiv.org/abs/2302.05601}, 
}

@misc{dong2017learningprunedeepneural,
      title={Learning to Prune Deep Neural Networks via Layer-wise Optimal Brain Surgeon}, 
      author={Xin Dong and Shangyu Chen and Sinno Jialin Pan},
      year={2017},
      eprint={1705.07565},
      archivePrefix={arXiv},
      primaryClass={cs.NE},
      url={https://arxiv.org/abs/1705.07565}, 
}

@inproceedings{NIPS1989_6c9882bb,
 author = {LeCun, Yann and Denker, John and Solla, Sara},
 booktitle = {Advances in Neural Information Processing Systems},
 editor = {D. Touretzky},
 pages = {},
 publisher = {Morgan-Kaufmann},
 title = {Optimal Brain Damage},
 url = {https://proceedings.neurips.cc/paper_files/paper/1989/file/6c9882bbac1c7093bd25041881277658-Paper.pdf},
 volume = {2},
 year = {1989}
}

@inproceedings{NIPS1992_303ed4c6,
 author = {Hassibi, Babak and Stork, David},
 booktitle = {Advances in Neural Information Processing Systems},
 editor = {S. Hanson and J. Cowan and C. Giles},
 pages = {},
 publisher = {Morgan-Kaufmann},
 title = {Second order derivatives for network pruning: Optimal Brain Surgeon},
 url = {https://proceedings.neurips.cc/paper_files/paper/1992/file/303ed4c69846ab36c2904d3ba8573050-Paper.pdf},
 volume = {5},
 year = {1992}
}

@misc{molchanov2019importanceestimationneuralnetwork,
      title={Importance Estimation for Neural Network Pruning}, 
      author={Pavlo Molchanov and Arun Mallya and Stephen Tyree and Iuri Frosio and Jan Kautz},
      year={2019},
      eprint={1906.10771},
      archivePrefix={arXiv},
      primaryClass={cs.LG},
      url={https://arxiv.org/abs/1906.10771}, 
}

@misc{han2016deepcompressioncompressingdeep,
      title={Deep Compression: Compressing Deep Neural Networks with Pruning, Trained Quantization and Huffman Coding}, 
      author={Song Han and Huizi Mao and William J. Dally},
      year={2016},
      eprint={1510.00149},
      archivePrefix={arXiv},
      primaryClass={cs.CV},
      url={https://arxiv.org/abs/1510.00149}, 
}

@misc{wan2024efficientlargelanguagemodels,
      title={Efficient Large Language Models: A Survey}, 
      author={Zhongwei Wan and Xin Wang and Che Liu and Samiul Alam and Yu Zheng and Jiachen Liu and Zhongnan Qu and Shen Yan and Yi Zhu and Quanlu Zhang and Mosharaf Chowdhury and Mi Zhang},
      year={2024},
      eprint={2312.03863},
      archivePrefix={arXiv},
      primaryClass={cs.CL},
      url={https://arxiv.org/abs/2312.03863}, 
}

@misc{wang2024modelcompressionefficientinference,
      title={Model Compression and Efficient Inference for Large Language Models: A Survey}, 
      author={Wenxiao Wang and Wei Chen and Yicong Luo and Yongliu Long and Zhengkai Lin and Liye Zhang and Binbin Lin and Deng Cai and Xiaofei He},
      year={2024},
      eprint={2402.09748},
      archivePrefix={arXiv},
      primaryClass={cs.CL},
      url={https://arxiv.org/abs/2402.09748}, 
}

@misc{kim2024shortenedllamadepthpruning,
      title={Shortened LLaMA: Depth Pruning for Large Language Models with Comparison of Retraining Methods}, 
      author={Bo-Kyeong Kim and Geonmin Kim and Tae-Ho Kim and Thibault Castells and Shinkook Choi and Junho Shin and Hyoung-Kyu Song},
      year={2024},
      eprint={2402.02834},
      archivePrefix={arXiv},
      primaryClass={cs.LG},
      url={https://arxiv.org/abs/2402.02834}, 
}

@misc{mallya2018packnetaddingmultipletasks,
      title={PackNet: Adding Multiple Tasks to a Single Network by Iterative Pruning}, 
      author={Arun Mallya and Svetlana Lazebnik},
      year={2018},
      eprint={1711.05769},
      archivePrefix={arXiv},
      primaryClass={cs.CV},
      url={https://arxiv.org/abs/1711.05769}
}

@misc{openai2024gpt4technicalreport,
      title={GPT-4 Technical Report}, 
      author={OpenAI and Josh Achiam and Steven Adler and Sandhini Agarwal and Lama Ahmad and others},
      year={2024},
      eprint={2303.08774},
      archivePrefix={arXiv},
      primaryClass={cs.CL},
      url={https://arxiv.org/abs/2303.08774}
}

@misc{vicuna2023,
    title = {Vicuna: An Open-Source Chatbot Impressing GPT-4 with 90\%* ChatGPT Quality},
    url = {https://lmsys.org/blog/2023-03-30-vicuna/},
    author = {Chiang, Wei-Lin and Li, Zhuohan and Lin, Zi and Sheng, Ying and Wu, Zhanghao and Zhang, Hao and Zheng, Lianmin and Zhuang, Siyuan and Zhuang, Yonghao and Gonzalez, Joseph E. and Stoica, Ion and Xing, Eric P.},
    month = {March},
    year = {2023}
}

@misc{workshop2023bloom176bparameteropenaccessmultilingual,
      title={BLOOM: A 176B-Parameter Open-Access Multilingual Language Model}, 
      author={BigScience Workshop and : and Teven Le Scao and Angela Fan and others},
      year={2023},
      eprint={2211.05100},
      archivePrefix={arXiv},
      primaryClass={cs.CL},
      url={https://arxiv.org/abs/2211.05100}
}

@misc{grattafiori2024llama3herdmodels,
      title={The Llama 3 Herd of Models}, 
      author={Aaron Grattafiori and Abhimanyu Dubey and others},
      year={2024},
      eprint={2407.21783},
      archivePrefix={arXiv},
      primaryClass={cs.AI},
      url={https://arxiv.org/abs/2407.21783}, 
}

@misc{molchanov2017pruningconvolutionalneuralnetworks,
      title={Pruning Convolutional Neural Networks for Resource Efficient Inference}, 
      author={Pavlo Molchanov and Stephen Tyree and Tero Karras and Timo Aila and Jan Kautz},
      year={2017},
      eprint={1611.06440},
      archivePrefix={arXiv},
      primaryClass={cs.LG},
      url={https://arxiv.org/abs/1611.06440}, 
}

@misc{wang2021convolutionalneuralnetworkpruning,
      title={Convolutional Neural Network Pruning with Structural Redundancy Reduction}, 
      author={Zi Wang and Chengcheng Li and Xiangyang Wang},
      year={2021},
      eprint={2104.03438},
      archivePrefix={arXiv},
      primaryClass={cs.CV},
      url={https://arxiv.org/abs/2104.03438}, 
}

@article{frankle2018lottery,
  title={The lottery ticket hypothesis: Finding sparse, trainable neural networks},
  author={Frankle, Jonathan and Carbin, Michael},
  journal={arXiv preprint arXiv:1803.03635},
  year={2018}
}

@misc{vaswani2023attentionneed,
      title={Attention Is All You Need}, 
      author={Ashish Vaswani and Noam Shazeer and Niki Parmar and Jakob Uszkoreit and Llion Jones and Aidan N. Gomez and Lukasz Kaiser and Illia Polosukhin},
      year={2023},
      eprint={1706.03762},
      archivePrefix={arXiv},
      primaryClass={cs.CL},
      url={https://arxiv.org/abs/1706.03762}, 
}

@misc{zhang2024loraprunestructuredpruningmeets,
      title={LoRAPrune: Structured Pruning Meets Low-Rank Parameter-Efficient Fine-Tuning}, 
      author={Mingyang Zhang and Hao Chen and Chunhua Shen and Zhen Yang and Linlin Ou and Xinyi Yu and Bohan Zhuang},
      year={2024},
      eprint={2305.18403},
      archivePrefix={arXiv},
      primaryClass={cs.LG},
      url={https://arxiv.org/abs/2305.18403}, 
}

@misc{zhou2024surveyefficientinferencelarge,
      title={A Survey on Efficient Inference for Large Language Models}, 
      author={Zixuan Zhou and Xuefei Ning and Ke Hong and Tianyu Fu and Jiaming Xu and Shiyao Li and Yuming Lou and Luning Wang and Zhihang Yuan and Xiuhong Li and Shengen Yan and Guohao Dai and Xiao-Ping Zhang and Yuhan Dong and Yu Wang},
      year={2024},
      eprint={2404.14294},
      archivePrefix={arXiv},
      primaryClass={cs.CL},
      url={https://arxiv.org/abs/2404.14294}, 
}

@article{jiang2023mistral,
  title={Mistral 7B},
  author={Jiang, Albert Q. and Sablayrolles, Alexandre and Mensch, Arthur and Bamford, Chris and Chaplot, Devendra Singh and de Las Casas, Diego and Bressand, Florian and Lengyel, Gianna and Lample, Guillaume and Saulnier, Lucile and Lavaud, Lélio Renard and Lachaux, Marie-Anne and Stock, Pierre and Le Scao, Teven and Lavril, Thibaut and Wang, Thomas and Lacroix, Timothée and El Sayed, William},
  journal={arXiv preprint arXiv:2310.06825},
  year={2023},
  url={https://arxiv.org/abs/2310.06825}
}

@misc{yin2025outlierweighedlayerwisesparsity,
      title={Outlier Weighed Layerwise Sparsity (OWL): A Missing Secret Sauce for Pruning LLMs to High Sparsity}, 
      author={Lu Yin and You Wu and Zhenyu Zhang and Cheng-Yu Hsieh and Yaqing Wang and Yiling Jia and Gen Li and Ajay Jaiswal and Mykola Pechenizkiy and Yi Liang and Michael Bendersky and Zhangyang Wang and Shiwei Liu},
      year={2025},
      eprint={2310.05175},
      archivePrefix={arXiv},
      primaryClass={cs.LG},
      url={https://arxiv.org/abs/2310.05175}, 
}

@misc{tang2025darwinlmevolutionarystructuredpruning,
      title={DarwinLM: Evolutionary Structured Pruning of Large Language Models}, 
      author={Shengkun Tang and Oliver Sieberling and Eldar Kurtic and Zhiqiang Shen and Dan Alistarh},
      year={2025},
      eprint={2502.07780},
      archivePrefix={arXiv},
      primaryClass={cs.LG},
      url={https://arxiv.org/abs/2502.07780}, 
}

@misc{kurtic2023ziplminferenceawarestructuredpruning,
      title={ZipLM: Inference-Aware Structured Pruning of Language Models}, 
      author={Eldar Kurtic and Elias Frantar and Dan Alistarh},
      year={2023},
      eprint={2302.04089},
      archivePrefix={arXiv},
      primaryClass={cs.LG},
      url={https://arxiv.org/abs/2302.04089}, 
}

@misc{deepseekai2025deepseekr1incentivizingreasoningcapability,
      title={DeepSeek-R1: Incentivizing Reasoning Capability in LLMs via Reinforcement Learning}, 
      author={DeepSeek-AI and Daya Guo and Dejian Yang and others},
      year={2025},
      eprint={2501.12948},
      archivePrefix={arXiv},
      primaryClass={cs.CL},
      url={https://arxiv.org/abs/2501.12948}, 
}

@misc{cobbe2021trainingverifierssolvemath,
      title={Training Verifiers to Solve Math Word Problems}, 
      author={Karl Cobbe and Vineet Kosaraju and Mohammad Bavarian and Mark Chen and Heewoo Jun and Lukasz Kaiser and Matthias Plappert and Jerry Tworek and Jacob Hilton and Reiichiro Nakano and Christopher Hesse and John Schulman},
      year={2021},
      eprint={2110.14168},
      archivePrefix={arXiv},
      primaryClass={cs.LG},
      url={https://arxiv.org/abs/2110.14168}, 
}

@misc{zhang2025reasoningmeetscompressionunderstanding,
      title={When Reasoning Meets Compression: Understanding the Effects of LLMs Compression on Large Reasoning Models}, 
      author={Nan Zhang and Eugene Kwek and Yusen Zhang and Ngoc-Hieu Nguyen and Prasenjit Mitra and Rui Zhang},
      year={2025},
      eprint={2504.02010},
      archivePrefix={arXiv},
      primaryClass={cs.LG},
      url={https://arxiv.org/abs/2504.02010}, 
}

@misc{sui2025stopoverthinkingsurveyefficient,
      title={Stop Overthinking: A Survey on Efficient Reasoning for Large Language Models}, 
      author={Yang Sui and Yu-Neng Chuang and Guanchu Wang and Jiamu Zhang and Tianyi Zhang and Jiayi Yuan and Hongyi Liu and Andrew Wen and Shaochen Zhong and Na Zou and Hanjie Chen and Xia Hu},
      year={2025},
      eprint={2503.16419},
      archivePrefix={arXiv},
      primaryClass={cs.CL},
      url={https://arxiv.org/abs/2503.16419}, 
}

@misc{das2024sizegradientsshapepruning,
      title={Beyond Size: How Gradients Shape Pruning Decisions in Large Language Models}, 
      author={Rocktim Jyoti Das and Mingjie Sun and Liqun Ma and Zhiqiang Shen},
      year={2024},
      eprint={2311.04902},
      archivePrefix={arXiv},
      primaryClass={cs.CL},
      url={https://arxiv.org/abs/2311.04902}, 
}

@misc{yang2025wandapruninglargelanguage,
      title={Wanda++: Pruning Large Language Models via Regional Gradients}, 
      author={Yifan Yang and Kai Zhen and Bhavana Ganesh and Aram Galstyan and Goeric Huybrechts and Markus Müller and Jonas M. Kübler and Rupak Vignesh Swaminathan and Athanasios Mouchtaris and Sravan Babu Bodapati and Nathan Susanj and Zheng Zhang and Jack FitzGerald and Abhishek Kumar},
      year={2025},
      eprint={2503.04992},
      archivePrefix={arXiv},
      primaryClass={cs.LG},
      url={https://arxiv.org/abs/2503.04992}, 
}

@misc{dong2024prunerzeroevolvingsymbolicpruning,
      title={Pruner-Zero: Evolving Symbolic Pruning Metric from scratch for Large Language Models}, 
      author={Peijie Dong and Lujun Li and Zhenheng Tang and Xiang Liu and Xinglin Pan and Qiang Wang and Xiaowen Chu},
      year={2024},
      eprint={2406.02924},
      archivePrefix={arXiv},
      primaryClass={cs.LG},
      url={https://arxiv.org/abs/2406.02924}, 
}

@misc{yang2025qwen3technicalreport,
      title={Qwen3 Technical Report}, 
      author={An Yang and Anfeng Li and Baosong Yang and Others},
      year={2025},
      eprint={2505.09388},
      archivePrefix={arXiv},
      primaryClass={cs.CL},
      url={https://arxiv.org/abs/2505.09388}, 
}

@misc{jin2020diseasedoespatienthave,
      title={What Disease does this Patient Have? A Large-scale Open Domain Question Answering Dataset from Medical Exams}, 
      author={Di Jin and Eileen Pan and Nassim Oufattole and Wei-Hung Weng and Hanyi Fang and Peter Szolovits},
      year={2020},
      eprint={2009.13081},
      archivePrefix={arXiv},
      primaryClass={cs.CL},
      url={https://arxiv.org/abs/2009.13081}, 
}

\appendix

\section{Appendix}
\label{sec:appendix}
\color{black}

\subsection{Related Works}
Pruning is a fundamental model-compression technique that removes redundant parameters through sparsity.
The pioneering OBD work~\cite{NIPS1989_6c9882bb} established a Taylor-series framework for importance estimation, followed by extensive CNN successes~\cite{han2016deepcompressioncompressingdeep,molchanov2017pruningconvolutionalneuralnetworks,wang2021convolutionalneuralnetworkpruning}.
With the rise of large language models, pruning has become crucial for efficient inference~\cite{wan2024efficientlargelanguagemodels,wang2024modelcompressionefficientinference,zhou2024surveyefficientinferencelarge}.
Because full retraining is prohibitive, recent work shifts to post-training pruning using lightweight calibration data.
According to sparsity patterns, methods are either unstructured, removing individual weights but requiring specialized kernels, or structured, pruning entire heads, channels, or layers for hardware-friendly acceleration~\cite{wan2024efficientlargelanguagemodels,wang2024modelcompressionefficientinference,ma2023llm}.
For structural pruning, estimating structural importance remains central: early CNN studies proposed summation-based aggregation~\cite{molchanov2019importanceestimationneuralnetwork}, and LLM-Pruner~\cite{ma2023llm} extended this idea to element-, vector-, and channel-level metrics.
Our work follows this line, focusing on post-training structured pruning for LLMs.

\subsection{Detailed Explanation, Equations and Algorithm}\label{sec:detailed_gisp}
\paragraph{Oneshot global pruning and GISP procedure.}  
Specifically, given a predefined number of iteration steps \( n \) and a target pruning ratio \( \rho \), GISP performs pruning iteratively using a small pruning ratio \( \rho_k \) at iteration \( k \). For each iteration, we conduct the one-shot pruning procedure, as detailed in \cref{sec:naive-one-shot-gp}. Additionally, to determine the pruning ratio \( \rho_k \) at each step, a linear scheduler is employed to linearly increase the pruning ratio until the target ratio \( \rho \) is reached in the final iteration. Notably, A key difference from one-shot pruning is that, in GISP, the importance scores in each iteration are computed based on the pruned model from the previous iteration. The detailed procedures are as follows:

\textbf{\circled{1} Pruning ratio scheduling.} Before pruning starts, we adopt a linear scheduler to obtain a set of target pruning ratios for each iteration step, where the pruning ratios are linearly increased. By doing so, the actual number of pruned weights remains the same for each iteration. 
Specifically, the linear scheduler can be formulated as:
\begin{equation} 
\small
\label{eq:sched}
    \rho_k = \begin{cases}
    \rho, & \text{if } n = 1, \\
    \frac{k-1}{n-1}\rho, & \text{if } n > 1, \quad k=1,2,\ldots,n.
\end{cases}
\end{equation}
where $\rho$ is the overall target ratio, $n$ marks the pre-defined iteration steps, $k$ is defined as the index of current step. 

\textbf{\circled{2} Structured weight importance estimation.} As illustrated in \cref{eq:score_glocal}, the importance score for global pruning is derived from the Taylor series expansion of the loss error with respect to weight perturbation caused by pruning. Since higher-order terms are significantly smaller in magnitude compared to the first-order term, we follow prior global pruning methods~\cite{molchanov2019importanceestimationneuralnetwork,molchanov2017pruningconvolutionalneuralnetworks} and adopt the first-order approximation as our importance score:   

\begin{equation} 
\small
\label{eq: imp}
\begin{aligned}
I_{W_i^j} 
= \left|\,
\frac{\partial \mathcal{L}(D)}{\partial W_i^j}\,W_i^j
\right|,
\end{aligned}
\end{equation}

Following that, to construct the structured importance score, we aggregate the element-wise importance scores by summing them within each structured group. These groups can correspond to attention heads in attention blocks or channels in the linear layers of feed-forward blocks.

\textbf{\circled{3} Global ranking and pruning.}
Once the structured importance scores are obtained, we perform a TopK process to select the weights with minimum values according to the pruning ratio of the current iteration step. 

Importantly, unlike prior global pruning methods for small models that directly rank importance scores across the entire network, we find that such an approach is less effective for LLMs due to significant structural differences between attention and MLP blocks. Specifically, attention blocks tend to exhibit substantially higher importance scores than MLP blocks, as illustrated in \cref{fig:ppl_its_analysis}(c), where the scores for attention layers are an order of magnitude greater. To mitigate this imbalance, we normalize the importance scores within attention and MLP blocks separately, ensuring they are brought to a comparable scale across the model.

In the end, the formulation of the normalized importance metrics is given by:
\begin{equation} 
\small
\label{eq: imp}
\begin{aligned}
I({\theta_{\text{head}}})
=\frac{I({\theta_{\text{head}}}) }{\lvert \theta_{\text{head}}\rvert}, I({\theta_{\text{channel}}}) 
=\frac{I({\theta_{\text{channel}}})}{\lvert \theta_{\text{channel}} \rvert}
\end{aligned}
\end{equation}
where $\lvert \theta \rvert$ marks the parameter count for the specified structure type. The formulation of the global ranking and the mask generation process can then be formulated as follows:
\begin{equation} 
\small
\label{eq: ranking}
    \begin{gathered} 
I(\theta_{k-1}) \;=\; [I(\theta_{k-1}^{\text{attn}}), I(\theta_{k-1}^{\text{mlp}})],
\tau_k\;=\;
\mathrm{TopK}_{\rho_k}
\Bigl(I(\theta_{k-1})\Bigr), 
\\
m_{k}
\;=\; 
\begin{cases}
1, & I(\theta_{k-1}) \;>\; \tau_k,\\
0, & \text{otherwise}.
\end{cases}
\end{gathered}
\end{equation}

where $\tau$ is the threshold for eliminating ongoing pruned modules, $m$ is the binary mask that applies to the model weight. Moreover, in each iteration, the importance score will be evaluated on the pruned model $\theta_{k-1}$ to reflect any cascading effect, distinguishing it from one-shot global pruning.

\color{black}

\subsection{Detailed Experimental Setup}\label{sec:detailed_exp_setup}

\begin{table*}[ht]
  \centering
  \caption{Detailed settings for CMQA calibration dataset and evaluation.}
  \label{tab:task_metrics}
    \resizebox{0.9\linewidth}{!}{
  \begin{tabular}{lcccc}
    \toprule
    Task       & Token Length & Actual Tokens & Accuracy & Template                                                                              \\
    \midrule
    BoolQ      & 410                     & 73 000                 & acc              & \texttt{\{passage\}\textbackslash nQuestion: \{question\}\textbackslash nAnswer:}           \\
    PIQA       & 160                     & 73 125                 & acc norm         & \texttt{Question: \{goal\}\textbackslash nAnswer:}                                         \\
    Hellaswag  & 144                     & 73 027                & acc norm         & \texttt{\{activity\_label\}: \{ctx\_a ctx\_b\}}                                             \\
    WinoGrande & 38                      & 73 117                 & acc              & \texttt{\{substituted\_sentence\_at\_bottomline\}}                                         \\
    ARC-E      & 92                      & 73 081                 & acc norm         & \texttt{Question: \{question\}\textbackslash nAnswer:}                                    \\
    ARC-C      & 112                     & 73 123               & acc norm         & \texttt{Question: \{question\}\textbackslash nAnswer:}                                    \\
    OBQA       & 43                      & 73 140                & acc norm         & \texttt{\{question\_stem\}}                                                                \\
    \bottomrule
  \end{tabular}
  }
\end{table*}

\begin{table*}[!htbp]
  \centering
  \caption{Detailed experimental hyper-parameters.}
  \label{tab:exp_settings}
    \resizebox{0.7\linewidth}{!}{
  \begin{tabular}{lcccc}
    \toprule           Model                                              & Random Seed & Precision   & Pruning Ratio/Iter \\
    \midrule
                               Llama2-7B, Llama3-8B, Mistral 0.3-7B               & 0         & bfloat16                 & 0.625\%       \\
                               Llama2-13B                                         & 0           & bfloat16   & 0.25\%        \\
    \bottomrule
  \end{tabular}
  }
\end{table*}

\begin{figure*}[!htbp]
    \centering
    \includegraphics[width=1\linewidth]{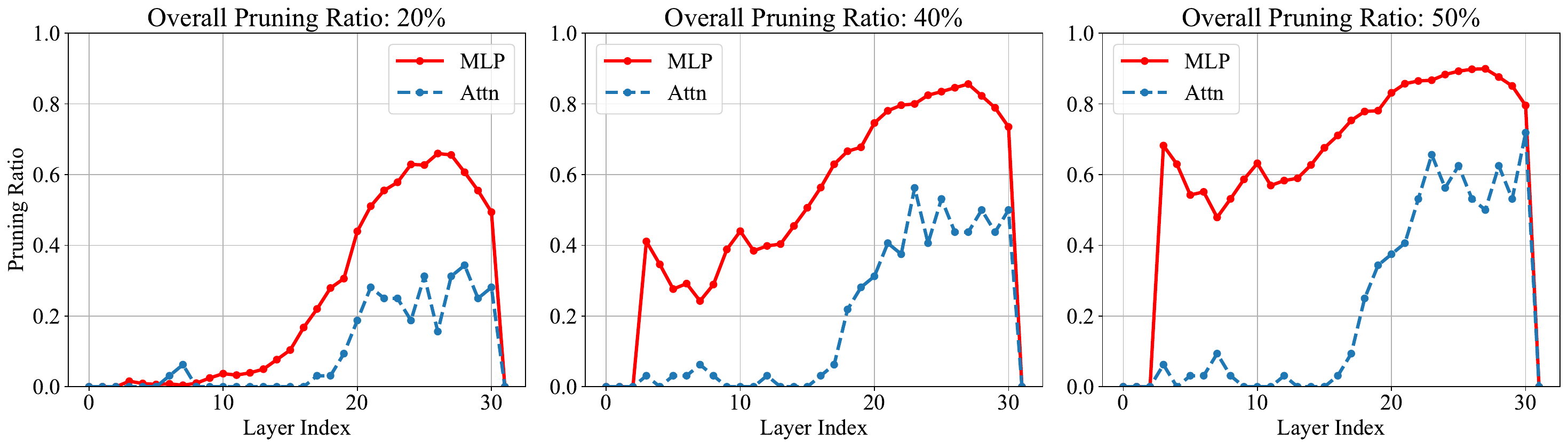}
    \caption{Visualization of the resulting model (Llama2-7B) in various overall pruning ratios from GISP.}\label{fig:GISP_sparsity}
\end{figure*}

\begin{figure*}[!htbp]
    \centering
    \includegraphics[width=1\linewidth]{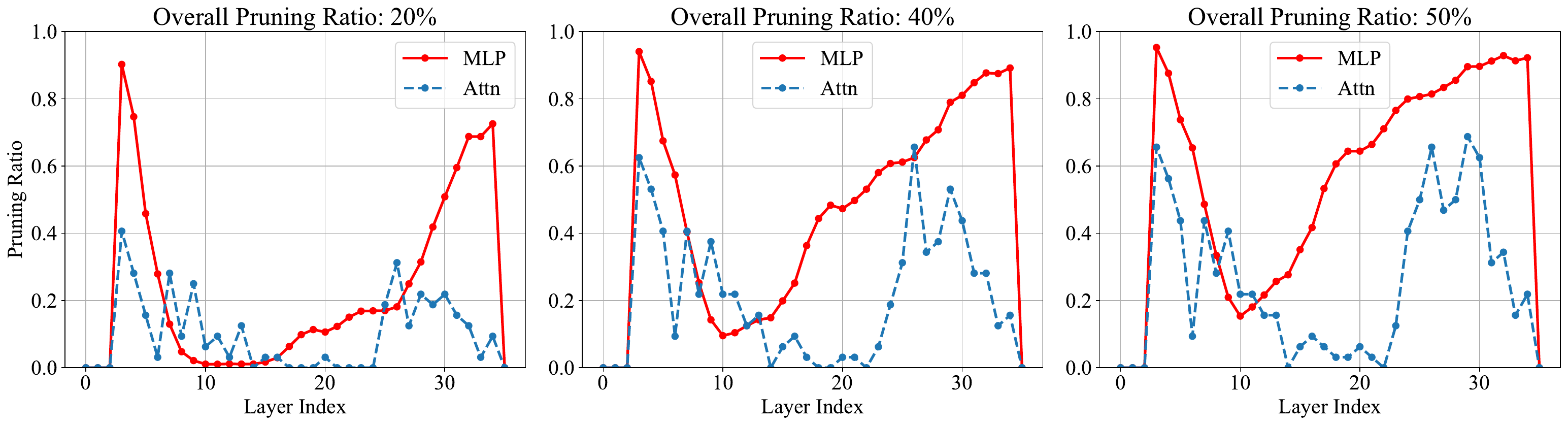}
    \caption{Visualization of the resulting model (Qwen3-8B) in various overall pruning ratios from GISP.}\label{fig:GISP_sparsity_qwen}
\end{figure*}
\paragraph{Experimental settings.} The detailed settings are listed in \cref{tab:task_metrics} and \cref{tab:exp_settings}. All baselines will receive the identical calibration dataset for pruning usage in each evaluation task. No re-training or recovery method is used, and only the pruning methods from baselines are evaluated for comparison. In addition, following the settings of Wanda-sp and LLM-pruner, we skip to prune the first 10\% of layers and the last layer. All experiments are conducted on a cloud computing server with an AMD EPYC 9554 CPU, 318.6 GB of memory, 400GB SSD, and one Nvidia H100 80GB GPU.

\paragraph{Text generation and zero-shot commonsense reasoning tasks.} Following the general setting, we use the C4 dataset as the calibration dataset for text generation tasks and zero-shot commonsense reasoning tasks, with 2000 samples, each having 256 token lengths. No template is used for this task. For the GSM8K task, we use both C4 and GSM8K as calibration and separately evaluate 8-shot accuracy.

\paragraph{Downstream commonsense reasoning tasks.} For the downstream commonsense reasoning tasks (CMQA), we use the CMQA training set as the calibration dataset with a total token budget of 512000 (matching previous C4 settings), which is then evenly distributed across each sub-task’s training split. To be specific, we include the gold answer (marked as positive labels) and all other options (marked as negative labels) for each sampled question from the training set, forming positive/negative pairs for margin evaluation. For individual tasks, we sample 2000 data points per task and set each task’s token‐length cap at the 99th percentile of these sampled data. The prompt templates follow the EleutherAI LM Harness pipeline conventions to ensure consistency between calibration and evaluation.  We report plain accuracy (acc) for fixed‐length tasks (e.g., true/false) and normalized accuracy (acc\_norm) for tasks with variable‐length answers, thus counteracting cumulative‐loss biases on longer sequences. 

\color{black}
\subsection{Comparison with SparseLLM}\label{sec:sparsellm_comparison}
SparseLLM, which was originally proposed for unstructured pruning \cite{bai2024sparsellmglobalpruningpretrained}, approximates Eq.~\ref{eqt:global_pruning} through a sequence of block-level reconstruction
constraints with iterative hidden-state updates to propagate pruning effects across blocks. To compare with SparseLLM, we build on the official SparseLLM codebase and implement a structured variant. All methods prune Llama2-7B using the same unified CMQA calibration set and are evaluated by the average accuracy over the 7-task CMQA suite, consistent with our experimental setup discussed in Section \ref{sec:exp} and \ref{sec:detailed_exp_setup}.

Following SparseLLM’s design, we use Wanda-SP as the local solver and apply SparseLLM’s optimization over all MLP components. As shown in Table~\ref{tab:sparsellm_cmqa}, SparseLLM brings only marginal changes over its local solver in this structured setting, whereas GISP consistently achieves higher accuracy, with larger gains at higher sparsity.

We conjecture that the limited gain of SparseLLM here may be due to:
(i) SparseLLM was originally designed for unstructured/semi-structured pruning. In contrast, under structured pruning, removing entire attention heads or MLP channels imposes structural constraints on the remaining network, which cannot be effectively adjusted through activation recomputation alone. Since SparseLLM relies on iteratively updating intermediate activations under a fixed pruning mask to reduce reconstruction error, its update mechanism becomes less effective when structured sparsity restricts the expressivity of the remaining feature space; and (ii) under structured pruning, performance can be highly sensitive to pruning decisions in attention blocks, while SparseLLM mainly improves the MLP/FFN part, so its benefits may be masked when attention heads are also removed.

\begin{table}[t]
\caption{Average CMQA accuracy (\%) on Llama2-7B acorss different pruning methods.}
\label{tab:sparsellm_cmqa}
\resizebox{1\linewidth}{!}{
\centering
\begin{tabular}{lcccc}
\toprule
Method & 20\% & 30\% & 40\% & 50\% \\
\midrule
Wanda-sp & 61.77 & 57.14 & 50.12 & 43.52 \\
SparseLLM (Wanda-sp solver) & 61.84 & 57.07 & 50.09 & 43.52 \\
GISP (ours) & \textbf{63.46} & \textbf{60.68} & \textbf{55.28} & \textbf{48.54} \\
\bottomrule
\end{tabular}
}
\end{table}
\color{black}

\begin{table}[t]
\centering
\caption{MedQA accuracy on Llama2-7B under various methods and pruning ratios.}
\label{tab:medqa}
\resizebox{0.8\linewidth}{!}{
\begin{tabular}{lcc}
\toprule
Method & Pruning ratio & MedQA ACC (\%) \\
\midrule
Dense        & 0\%  & 32.05 \\
Wanda-sp     & 20\% & 21.92 \\
GISP (ours)  & 20\% & \textbf{31.26} \\
GISP (ours)  & 30\% & \textbf{29.77} \\
GISP (ours)  & 40\% & \textbf{26.24} \\
\bottomrule
\end{tabular}
}
\end{table}

\color{black}
\subsection{Performance of GISP on domain-specific QA tasks}\label{sec:app_medqa}
To assess whether the task-specific benefits of GISP extend beyond commonsense and mathematical reasoning, we also evaluate MedQA, a medical exam-style multiple-choice benchmark requiring specialized biomedical knowledge. 

Following the same structured pruning protocol, we construct a MedQA calibration set with a 512K-token budget and prune Llama2-7B accordingly. As shown in Table~\ref{tab:medqa}, GISP degrades much more gracefully with increasing sparsity, indicating stronger robustness to pruning. More importantly, its performance at 40\% sparsity already exceeds that of Wanda-sp at 20\% sparsity, highlighting the effectiveness of GISP in maintaining domain knowledge via leveraging task-specific information from the calibration dataset even in substantially sparser settings.

\subsection{Detailed Analysis Result of Robustness}\label{sec:app_detailed_robust}
In this section, we report the exact numerical results corresponding to Sec.~\ref{sec:calib_robust}.
\paragraph{Full results under calibration-budget variation.}
We subsample the default 512K-token CMQA calibration budget to 1\%, 20\%, 40\%, 60\%, 80\%, and 100\%, while keeping all other settings fixed and rerunning GISP under each budget. We use 0 as the random seed, consistent with Sec.\ref{sec:exp_main_result} and Sec.~\ref{sec:detailed_exp_setup}.
Table~\ref{tab:calib_budget_full} reports the resulting CMQA accuracy under four target sparsities.


\begin{table}[t]
\caption{CMQA average accuracy (\%) of GISP on Llama2-7B under different calibration budgets.}
\label{tab:calib_budget_full}
\resizebox{1\linewidth}{!}{
\centering
\begin{tabular}{lcccc}
\toprule
Token budget & 20\% & 30\% & 40\% & 50\% \\
\midrule
5,120 (1\%)       & 63.86 & 60.46 & 55.83 & 49.20 \\
102,400 (20\%)    & 64.19 & 61.81 & 56.85 & 50.64 \\
204,800 (40\%)    & 64.00 & 61.32 & 56.70 & 50.40 \\
307,200 (60\%)    & 63.93 & 61.45 & 56.05 & 49.82 \\
409,600 (80\%)    & 63.93 & 61.00 & 56.29 & 49.29 \\
512,000 (100\%)   & 63.46 & 60.68 & 55.28 & 48.54 \\
\midrule
Std.              & 0.24 & 0.50 & 0.58 & 0.79 \\
\bottomrule
\end{tabular}
}
\end{table}

\paragraph{Trajectory statistics under reduced calibration budgets.}
We further rerun the full iterative pruning procedure using random seed 0 to subsample under five reduced calibration budgets (1\%, 20\%, 40\%, 60\%, and 80\% of the default budget), and evaluate 10 sparsity points from 5\% to 50\%.
Table~\ref{tab:trajectory_budget_stability} reports the average accuracy and standard deviation across these five budget settings. Fig.~\ref{fig:calib_robust} in the main text visualizes these trajectory statistics.

These results suggest that GISP is reasonably robust to calibration-budget limitations in this setting, and that moderate reductions in calibration data do not substantially compromise performance and trajectory stability. Note that the extra fine-granularity intermediate evaluation process may cause the random number to change and thus obtain slightly different pruning results.

\begin{table}[t]
\centering
\caption{Stability of the iterative pruning trajectory under calibration-budget variation.}
\label{tab:trajectory_budget_stability}
\resizebox{0.8\linewidth}{!}{
\begin{tabular}{ccc}
\toprule
Pruning ratio & AVG ACC (\%) & STD (\%) \\
\midrule
5\%  & 66.14 & 0.13 \\
10\% & 65.62 & 0.11 \\
15\% & 64.85 & 0.18 \\
20\% & 63.88 & 0.20 \\
25\% & 62.50 & 0.39 \\
30\% & 61.26 & 0.49 \\
35\% & 59.30 & 0.75 \\
40\% & 56.35 & 0.41 \\
45\% & 53.26 & 0.50 \\
50\% & 49.94 & 0.69 \\
\bottomrule
\end{tabular}
}
\end{table}

\color{black}

\subsection{Detailed Downstream CMQA Accuracy}\label{sec:detailed_downstream}

We provided detailed downstream task accuracy evaluations at \cref{tab:llama2-7b}, \cref{tab:llama3-8b}, \cref{tab:mistral0.3-7b}, and \cref{tab:llama2-13b}. We present our key observations of these detailed evaluations as follows.

\textit{(1)} On downstream tasks, GISP consistently achieves higher accuracy across all models and pruning ratios, with particularly strong gains at higher pruning levels, \textit{indicating its strength as a task-specific pruner}. For example, on the BoolQ task, GISP holds a 6–20\% accuracy lead over the best baseline at every pruning ratio. Moreover, at 30–40\% pruning ratio, GISP’s accuracy remains close to the dense model—for instance, 80.00\% (ours) vs. 80.55\% (dense) on Llama2-13B at 40\% pruning ratio—while other methods begin to lose accuracy even at lower pruning ratios.

\textit{(2)} For local pruning methods, downstream performance remains similar to—or even lower than—their zero-shot performance, suggesting that while local pruning preserves general knowledge, it lacks task-specific optimization.

\subsection{Extended Ablation of Calibration Dataset and Task-specific Loss Target}

We provided an extended ablation of using different calibration datasets and the effectiveness of GISP as a task-specific pruner with task-specific loss target at \cref{tab:calibration_full}.

\textit{(1) GISP is inherently task-specific.} When we switch from C4 to CMQA for calibration, GISP gains significant accuracy improvements at every pruning ratio. In contrast, all baselines show no uplift (and sometimes even regress), reflecting their general-purpose properties with no sensitivity to task information and highlighting GISP’s task-specific capability.

\textit{(2) Effectiveness of GISP as a task-specific pruner with task-specific loss target.} Incorporating the margin-based loss provides consistent accuracy gains across nearly every task and pruning ratio, showing the necessity and effectiveness of GISP's design to accommodate various task-specific loss targets.

\subsection{Visualization of the pruned model} 
Figure \ref{fig:GISP_sparsity} and \ref{fig:GISP_sparsity_qwen} provide the layer-wise pruning ratio distribution of various target pruning ratios of GISP on attention blocks and MLP blocks of the Llama2-7B and Qwen3-8B models, respectively. We present our key observations of the generated pruned model as follows, aiming to provide insight for further works, such as LLMs architecture search, design, and explanation:

\textit{(1) Layer-wise sparsity varies significantly for both attention and MLP components.} Both the MLP and attention layers exhibit a similar trend of increasing pruning ratios from early to late layers, suggesting that earlier layers are more critical to model performance than later ones.

\textit{(2) MLP layers are more redundant than attention layers.} First, the pruning ratio of MLP layers is consistently higher than that of attention layers across all layers. 

\color{black}
\textit{(3) Different model families exhibit distinct redundancy layouts.} For example, in Llama2, redundancy is more concentrated in later layers, whereas Qwen3 shows a characteristic “high–low–high” (concave) layer-wise sparsity pattern. These differences indicate that GISP is sensitive to architectural characteristics and discovers model-specific sparsity structures.

Overall, these analyses support that GISP consistently removes redundant components while preserving more critical structures such as middle attention layers. At the same time, the discovered sparsity profiles meaningfully vary with model architecture, further demonstrating that GISP performs structured, data-driven pruning rather than heuristic removal.

\color{black}

\color{black}
\subsection{Efficiency Analysis}
\paragraph{Pruning procedure efficiency analysis.}
The runtime overhead of deploying GISP is discussed in Section \ref{sec:prune-once-for-all} and summarized in Table \ref{tab:pruning_time}.

We additionally report the peak GPU memory during pruning on Llama2-7B under the same environment (batch size and calibration setting) in Table~\ref{tab:pruning_mem}. As expected and claimed in limitation section, GISP uses higher peak memory than local forward-only baselines, as it requires a backward pass for gradient computation.

\begin{table}[t]
\caption{Peak GPU memory during pruning on Llama2-7B.}
\centering
\small
\begin{tabular}{l r}
\toprule
Method & Peak memory (MiB) \\
\midrule
GISP & 43576 \\
Wanda & 22128 \\
FLAP & 21656 \\
OWL  & 26528 \\
\bottomrule
\end{tabular}
\label{tab:pruning_mem}
\end{table}

\paragraph{Pruned model efficiency analysis.}
We evaluate the inference throughput and memory footprint of the pruned Llama2-7B model on an NVIDIA A6000 GPU with batch size $=1$ and report time-to-first-token (TTFT), prefill/decode throughput, related speedup and memory cost in Table~\ref{tab:throughput} and ~\ref{tab:infer_mem}. GISP constantly reduces latency, improves both decoding and prefilling speedup and reduce memory footprint compared with the dense model.

\begin{table}[t]
\caption{Inference efficiency of pruned subnetworks.}
\label{tab:throughput}
\resizebox{1\linewidth}{!}{
\centering
\begin{tabular}{lccccc}
\toprule
Pruning Ratio & TTFT (s) & Prefill (t/s) & Decode (t/s) & Prefill Inc. & Decode Inc. \\
\midrule
Dense & 0.105 & 5635.4 & 38.0 & 1.00x & 1.00x\\
20\%  & 0.090 & 6833.6 & 41.4 & 1.21x & 1.09x\\
30\%  & 0.082 & 7704.7 & 43.2 & 1.37x & 1.14x\\
40\%  & 0.076 & 8396.9 & 43.7 & 1.49x & 1.15x\\
50\%  & 0.069 & 9562.7 & 43.7 & 1.70x & 1.15x\\
\bottomrule
\end{tabular}
}
\end{table}

\begin{table}[t]
\centering
\caption{Inference memory footprint.}
\small
\begin{tabular}{c r r}
\toprule
Sparsity & Memory (GB) & Reduction (\%) \\
\midrule
Dense & 14.05 & -- \\
20\%  & 11.58 & 17.6 \\
30\%  & 10.27 & 26.9 \\
40\%  &  9.07 & 35.4 \\
50\%  &  7.73 & 45.0 \\
\bottomrule
\end{tabular}
\label{tab:infer_mem}
\end{table}

\color{black}
\subsection{Practical Impact of the GISP: Model saving and on-the-fly adaptation}
Thanks to the once-for-all property, GISP can produce a spectrum of models pruned to different pruning ratios to a target ratio $\rho$. To exploit this without extra storage overhead, we record only the indices of channels or heads removed at each iteration—orders of magnitude smaller than element-wise masks. Once pruning is complete, any intermediate pruned model can be reconstructed simply by reapplying the saved indices. \textcolor{black}{For example, after applying GISP to the Llama2-7B model, users only need to store one set of weights with the corresponding masks (approximately 318KB per mask, which is ~0.002\% of the model weight for Llama2-7b), rather than maintaining numerous separate models.}

This enables on-the-fly adaptation: by running GISP as a preprocessing step to capture the pruning schedule, users can dynamically deploy the most suitable pruned model according to available compute resources and deployment environments. 

\color{black}
\subsection{GISP Under Higher Pruning Ratio}

\begin{table}[t]
\caption{Performance of pruned model under 60\% pruning ratio on Llama2-7B.}
\centering
\small
\begin{tabular}{lcc}
\toprule
Method & Pruning ratio & 0shot CMQA ACC (\%) \\
\midrule
Dense & 0\% & 66.68 \\
GISP & 60\% & 42.17 \\
Wanda & 60\% & 37.77 \\
FLAP & 60\% &  37.98 \\
OWL  & 60\% & 37.81 \\
\bottomrule
\end{tabular}
\label{tab:pruning_higher_ratio}
\end{table}
To further demonstrate the performance advantage of GISP in higher sparsity, we add an additional pruning experiment on Llama2-7B, targeting 60\% sparsity, for reference. The results in Table \ref{tab:pruning_higher_ratio} show the expected trend: sharp performance drops for all baseline methods, while GISP remains competitive and degrades more gracefully.
\color{black}
\subsection{LLM Usage}  
In accordance with the ARR AI Writing/Coding Assistance Policy, we disclose that LLM-based tools (e.g., ChatGPT) were used solely to aid in polishing the writing and improving the clarity of exposition. They were not used for research ideation, experimental design, data analysis, or other substantive contributions. All scientific content, results, and conclusions are the responsibility of the authors.

\begin{table*}[htbp]
  \centering
  \caption{Llama 2-7B Downstream CMQA Accuracy under Different Pruning Methods.}
  \label{tab:llama2-7b}

\resizebox{0.9\linewidth}{!}{
  \begin{tabular}{l c c c c c c c c c}
    \toprule
    Method        & Pruning Ratio & BoolQ  & PIQA   & Hellaswag & WinoGrande & ARC-E & ARC-C & OBQA   & AVG  \\
    \midrule
    Dense         & 0\%     & 77.71  & 79.05  & 76.00     & 68.98      & 74.54    & 46.25         & 44.20  & 66.68  \\
    \midrule
    \multirow{4}{*}{Wanda-sp}
                  & 20\%    & 65.84  & 78.73  & 71.07     & 62.75      & 69.32    & 43.09         & 41.60  & 61.77  \\
                  & 30\%    & 62.35  & 75.73  & 63.44     & 57.85      & 63.30    & 38.48         & 38.80  & 57.14  \\
                  & 40\%    & 61.38  & 73.39  & 44.69     & 50.91      & 53.87    & 30.38         & 36.20  & 50.12  \\
                  & 50\%    & 58.17  & 65.29  & 34.60     & 50.75      & 40.82    & 24.57         & 30.40  & 43.52  \\
    \midrule
    \multirow{4}{*}{LLM-Pruner}
                  & 20\%    & 64.37  & 71.44  & 48.49     & 57.85      & 55.85    & 28.24         & 30.40  & 50.95  \\
                  & 30\%    & 60.31  & 60.94  & 31.31     & 50.67      & 38.80    & 20.73         & 26.80  & 41.37  \\
                  & 40\%    & 60.55  & 55.17  & 28.71     & 49.41      & 33.12    & 20.14         & 26.80  & 39.13  \\
                  & 50\%    & 60.92  & 53.70  & 28.07     & 50.12      & 31.57    & 20.73         & 25.20  & 38.62  \\
    \midrule
    \multirow{4}{*}{FLAP}
                  & 20\%    & 67.16  & 77.48  & 70.64     & 62.35      & 66.54    & 42.49         & 42.20  & 61.27  \\
                  & 30\%    & 62.87  & 75.24  & 63.47     & 57.85      & 61.32    & 38.74         & 38.80  & 56.90  \\
                  & 40\%    & 61.65  & 72.03  & 53.86     & 54.22      & 55.35    & 36.95         & 37.00  & 53.01  \\
                  & 50\%    & 59.45  & 68.28  & 42.58     & 53.12      & 48.53    & 30.72         & 32.20  & 47.84  \\
    \midrule
    \multirow{4}{*}{ShortGPT}
                  & 20\%    & 62.17  & 70.18  & 62.73     & 65.82      & 55.93    & 36.18         & 37.20  & 55.75  \\
                  & 30\%    & 62.20  & 63.38  & 50.80     & 62.98      & 45.08    & 34.22         & 31.40  & 50.01  \\
                  & 40\%    & 62.17  & 57.83  & 41.16     & 58.09      & 37.08    & 30.12         & 31.00  & 45.35  \\
                  & 50\%    & 62.17  & 52.61  & 33.35     & 56.91      & 31.73    & 26.71         & 28.80  & 41.75  \\
    \midrule
    \multirow{4}{*}{OWL}
                  & 20\%    & 67.09  & 78.67  & 71.87     & 66.14      & 69.87    & 43.43         & 41.40  & 62.64  \\
                  & 30\%    & 64.04  & 76.66  & 66.54     & 58.33      & 64.44    & 38.91         & 39.40  & 58.33  \\
                  & 40\%    & 62.14  & 74.05  & 47.62     & 52.49      & 55.01    & 30.80         & 38.40  & 51.50  \\
                  & 50\%    & 61.25  & 66.16  & 35.41     & 51.62      & 44.15    & 24.91         & 30.20  & 44.82  \\
    \midrule
    \multirow{4}{*}{GISP (ours)}
                  & 20\%    & 77.77  & 75.30  & 70.71     & 69.14      & 69.65    & 41.64         & 40.00  & \textbf{63.46}  \\
                  & 30\%    & 77.19  & 72.85  & 64.19     & 65.35      & 65.24    & 40.53         & 39.40  & \textbf{60.68}  \\
                  & 40\%    & 70.55  & 68.88  & 53.68     & 62.51      & 57.74    & 35.41         & 38.20  & \textbf{55.28}  \\
                  & 50\%    & 65.29  & 64.09  & 41.27     & 56.27      & 49.71    & 29.18         & 34.00  & \textbf{48.54}  \\
    \bottomrule
  \end{tabular}
}
\end{table*}

\begin{table*}[htbp]
  \centering
  \caption{Llama 3-8B Downstream CMQA Accuracy under Different Pruning Methods.}
  \label{tab:llama3-8b}
  \resizebox{0.9\linewidth}{!}{
  \begin{tabular}{l c c c c c c c c c}
    \toprule
    Method        & Pruning Ratio & BoolQ  & PIQA   & Hellaswag & WinoGrande & ARC-E & ARC-C & OBQA   & AVG    \\
    \midrule
    Dense         & 0\%           & 81.28  & 80.79  & 79.13     & 72.61      & 77.69 & 53.41 & 45.00  & 69.99             \\
    \midrule
    \multirow{4}{*}{Wanda-sp}
                  & 20\%          & 59.42  & 77.31  & 58.77     & 59.67      & 67.68 & 39.68 & 39.60  & 57.45             \\
                  & 30\%          & 61.28  & 74.43  & 46.62     & 54.46      & 57.66 & 32.94 & 36.80  & 52.03             \\
                  & 40\%          & 62.17  & 63.93  & 33.15     & 52.09      & 42.93 & 23.98 & 27.00  & 43.61             \\
                  & 50\%          & 58.75  & 60.88  & 31.87     & 50.67      & 37.84 & 23.04 & 26.20  & 41.32             \\
    \midrule
    \multirow{4}{*}{LLM-Pruner}
                  & 20\%          & 67.68  & 75.03  & 57.76     & 60.77      & 61.28 & 37.03 & 36.00  & 56.51             \\
                  & 30\%          & 60.86  & 66.38  & 41.35     & 54.54      & 51.35 & 27.13 & 30.60  & 47.46             \\
                  & 40\%          & 57.31  & 60.61  & 33.68     & 51.46      & 39.94 & 22.53 & 27.20  & 41.82             \\
                  & 50\%          & 52.63  & 56.37  & 31.08     & 50.43      & 36.07 & 22.53 & 28.60  & 39.67             \\
    \midrule
    \multirow{4}{*}{ShortGPT}
                  & 20\%          & 65.02  & 71.00  & 64.61     & 70.88      & 56.65 & 42.41 & 33.20  & 57.68             \\
                  & 30\%          & 51.68  & 60.72  & 33.44     & 58.48      & 39.27 & 30.55 & 30.60  & 43.53             \\
                  & 40\%          & 58.62  & 60.45  & 37.76     & 52.96      & 35.23 & 29.95 & 28.60  & 43.37             \\
                  & 50\%          & 60.86  & 55.39  & 29.38     & 54.54      & 29.71 & 28.84 & 29.60  & 41.19             \\
    \midrule
    \multirow{4}{*}{OWL}
                  & 20\%          & 64.74  & 77.97  & 61.82     & 62.83      & 70.58 & 41.13 & 40.60  & 59.95             \\
                  & 30\%          & 62.84  & 73.29  & 47.51     & 55.88      & 56.90 & 32.25 & 37.00  & 52.24             \\
                  & 40\%          & 62.11  & 65.18  & 34.24     & 52.09      & 45.58 & 24.66 & 30.20  & 44.87             \\
                  & 50\%          & 60.09  & 60.12  & 31.37     & 52.72      & 38.76 & 22.95 & 27.00  & 41.86             \\
    \midrule
    \multirow{4}{*}{GISP (ours)}
                  & 20\%          & 79.11  & 76.50  & 70.16     & 71.43      & 70.66 & 47.10 & 42.00  & \textbf{65.28}             \\
                  & 30\%          & 78.04  & 71.00  & 59.24     & 69.93      & 62.46 & 40.53 & 36.40  & \textbf{59.66}             \\
                  & 40\%          & 72.69  & 67.25  & 47.58     & 66.14      & 54.59 & 34.13 & 32.20  & \textbf{53.51}             \\
                  & 50\%          & 66.48  & 62.19  & 36.07     & 55.72      & 42.34 & 27.99 & 29.00  & \textbf{45.68}             \\
    \bottomrule
  \end{tabular}
  }
\end{table*}

\begin{table*}[htbp]
  \centering
  \caption{Mistral 0.3-7B Downstream CMQA Accuracy under Different Pruning Methods.}
  \label{tab:mistral0.3-7b}
  \resizebox{0.9\linewidth}{!}{
  \begin{tabular}{l c c c c c c c c c}
    \toprule
    Method        & Pruning Ratio & BoolQ  & PIQA   & Hellaswag & WinoGrande & ARC-E & ARC-C & OBQA   & AVG    \\
    \midrule
    Dense         & 0\%      & 82.08  & 82.21  & 80.44     & 73.88      & 78.24 & 52.22 & 44.20  & 70.47  \\
    \midrule
    \multirow{4}{*}{Wanda-sp}
                  & 20\%     & 68.72  & 80.52  & 72.73     & 66.77      & 74.79 & 44.03 & 43.20  & 64.39  \\
                  & 30\%     & 57.92  & 78.89  & 63.17     & 58.56      & 68.81 & 37.54 & 42.80  & 58.24  \\
                  & 40\%     & 53.39  & 76.22  & 50.92     & 55.88      & 58.21 & 31.40 & 37.20  & 51.89  \\
                  & 50\%     & 58.04  & 67.03  & 36.53     & 49.96      & 45.66 & 24.23 & 29.20  & 44.38  \\
    \midrule
    \multirow{4}{*}{ShortGPT}
                  & 20\%     & 69.36  & 72.31  & 64.71     & 68.51      & 58.63 & 39.16 & 31.60  & 57.75  \\
                  & 30\%     & 42.29  & 58.60  & 34.54     & 57.70      & 31.27 & 31.91 & 31.40  & 41.10  \\
                  & 40\%     & 53.00  & 53.10  & 26.40     & 55.80      & 30.47 & 30.80 & 28.20  & 39.68  \\
                  & 50\%     & 52.97  & 50.16  & 24.96     & 53.75      & 30.13 & 30.72 & 28.40  & 38.73  \\
    \midrule
    \multirow{4}{*}{OWL}
                  & 20\%     & 68.65  & 80.69  & 74.52     & 70.01      & 75.76 & 46.25 & 45.20  & 65.87  \\
                  & 30\%     & 52.60  & 79.05  & 66.04     & 61.25      & 69.40 & 39.25 & 42.20  & 58.54  \\
                  & 40\%     & 57.06  & 78.07  & 53.76     & 55.56      & 62.79 & 33.87 & 39.40  & 54.36  \\
                  & 50\%     & 61.19  & 69.70  & 38.66     & 50.99      & 46.38 & 26.11 & 29.60  & 46.09  \\
    \midrule
    \multirow{4}{*}{GISP (ours)}
                  & 20\%     & 80.52  & 78.40  & 73.55     & 73.16      & 74.54 & 47.01 & 39.00  & \textbf{66.60}  \\
                  & 30\%     & 79.79  & 76.06  & 65.69     & 70.01      & 71.68 & 41.55 & 39.60  & \textbf{63.48}  \\
                  & 40\%     & 77.00  & 71.71  & 54.57     & 65.04      & 64.60 & 37.97 & 37.20  & \textbf{58.30}  \\
                  & 50\%     & 70.21  & 62.79  & 40.36     & 56.75      & 55.26 & 32.17 & 31.00  & \textbf{49.79}  \\
    \bottomrule
  \end{tabular}
  }
\end{table*}
\begin{table*}[htbp]
  \centering
  \caption{Llama 2-13B Downstream CMQA Accuracy under Different Pruning Methods.}
  \label{tab:llama2-13b}
  \resizebox{0.9\linewidth}{!}{
  \begin{tabular}{l c c c c c c c c c}
    \toprule
    Method        & Pruning Ratio & BoolQ  & PIQA   & Hellaswag & WinoGrande & ARC-E & ARC-C & OBQA   & AVG    \\
    \midrule
    Dense         & 0\%      & 80.55  & 80.52  & 79.39     & 72.22      & 77.48 & 48.98 & 45.20  & 69.19  \\
    \midrule
    \multirow{4}{*}{Wanda-sp}
                  & 20\%     & 73.09  & 79.98  & 76.04     & 67.09      & 75.46 & 47.87 & 44.60  & 66.30  \\
                  & 30\%     & 70.83  & 78.94  & 68.68     & 62.83      & 71.30 & 44.03 & 42.20  & 62.69  \\
                  & 40\%     & 64.16  & 78.02  & 62.27     & 59.27      & 67.26 & 41.21 & 41.60  & 59.11  \\
                  & 50\%     & 62.14  & 71.98  & 48.91     & 53.35      & 55.13 & 32.68 & 37.00  & 51.60  \\
    \midrule
    \multirow{4}{*}{LLM-Pruner}
                  & 20\%     & 66.51  & 74.92  & 58.47     & 62.67      & 66.41 & 35.92 & 36.20  & 57.30  \\
                  & 30\%     & 61.77  & 67.19  & 36.64     & 51.54      & 51.01 & 25.26 & 30.40  & 46.26  \\
                  & 40\%     & 58.93  & 61.43  & 31.23     & 49.25      & 42.51 & 22.78 & 27.80  & 41.99  \\
                  & 50\%     & 61.96  & 58.11  & 28.81     & 51.70      & 36.78 & 22.27 & 26.80  & 40.92  \\
    \midrule
    \multirow{4}{*}{FLAP}
                  & 20\%     & 73.00  & 79.92  & 72.26     & 66.85      & 73.74 & 45.82 & 43.40  & 65.00  \\
                  & 30\%     & 69.94  & 78.35  & 67.79     & 65.11      & 71.72 & 44.88 & 45.20  & 63.28  \\
                  & 40\%     & 66.21  & 76.82  & 64.74     & 63.14      & 65.53 & 42.58 & 40.20  & 59.89  \\
                  & 50\%     & 64.16  & 74.48  & 57.94     & 57.77      & 60.82 & 38.65 & 40.20  & 56.29  \\
    \midrule
    \multirow{4}{*}{ShortGPT}
                  & 20\%     & 61.80  & 74.16  & 70.62     & 70.17      & 65.87 & 42.49 & 40.80  & 60.84  \\
                  & 30\%     & 61.53  & 69.80  & 64.57     & 69.53      & 55.47 & 39.33 & 37.80  & 56.86  \\
                  & 40\%     & 44.98  & 65.02  & 33.19     & 51.66      & 66.46 & 46.17 & 33.60  & 48.73  \\
                  & 50\%     & 62.17  & 52.61  & 33.35     & 56.91      & 31.73 & 26.71 & 28.80  & 41.75  \\
    \midrule
    \multirow{4}{*}{OWL}
                  & 20\%     & 76.48  & 80.30  & 77.19     & 68.35      & 74.71 & 46.84 & 45.00  & 66.98  \\
                  & 30\%     & 69.82  & 78.40  & 73.28     & 64.09      & 71.76 & 43.17 & 42.40  & 63.27  \\
                  & 40\%     & 69.97  & 77.69  & 66.97     & 60.62      & 67.51 & 42.92 & 40.20  & 60.84  \\
                  & 50\%     & 63.46  & 73.18  & 55.15     & 54.78      & 57.58 & 37.29 & 37.80  & 54.17  \\
    \midrule
    \multirow{4}{*}{GISP (ours)}
                  & 20\%     & 80.28  & 79.00  & 76.83     & 72.22      & 75.17 & 45.99 & 43.80  & \textbf{67.61}  \\
                  & 30\%     & 81.16  & 76.77  & 72.87     & 71.59      & 72.94 & 45.73 & 41.80  & \textbf{66.12}  \\
                  & 40\%     & 80.00  & 73.83  & 65.79     & 70.09      & 70.16 & 43.52 & 40.00  & \textbf{63.34}  \\
                  & 50\%     & 72.97  & 69.91  & 55.15     & 65.59      & 63.09 & 38.23 & 37.60  & \textbf{57.50}  \\
    \bottomrule
  \end{tabular}
    }
\end{table*}

\begin{table*}[htbp]
  \centering
  \caption{CMQA Accuracy on All Seven Tasks under Different Calibration Datasets and Pruning Ratios.}
  \label{tab:calibration_full}
    \resizebox{1\linewidth}{!}{
  \begin{tabular}{llccccccccc}
    \toprule
    Method        & Calibration Dataset   & Pruning Ratio & BoolQ & PIQA  & Hellaswag & WinoGrande & ARC-E & ARC-C & OBQA  & AVG   \\
    \midrule
    \multirow{8}{*}{Wanda-sp} 
                 & \multirow{4}{*}{C4}   & 20\%     & 74.07 & 79.33 & 77.94     & 70.48      & 72.77  & 45.14  & 44.80 & 66.36 \\
                 &                       & 30\%     & 67.86 & 78.02 & 74.65     & 68.11      & 69.57  & 43.52  & 44.20 & 63.70 \\
                 &                       & 40\%     & 64.40 & 77.31 & 68.36     & 61.40      & 57.41  & 37.97  & 40.80 & 58.24 \\
                 &                       & 50\%     & 62.63 & 72.31 & 58.69     & 55.96      & 48.32  & 31.23  & 36.60 & 52.25 \\
    \cmidrule(l){2-11}
                 & \multirow{4}{*}{CMQA} & 20\%     & 73.09 & 79.98 & 76.04     & 67.09      & 75.46  & 47.87  & 44.60 & 66.30 \\
                 &                       & 30\%     & 70.83 & 78.94 & 68.68     & 62.83      & 71.30  & 44.03  & 42.20 & 62.69 \\
                 &                       & 40\%     & 64.16 & 78.02 & 62.27     & 59.27      & 67.26  & 41.21  & 41.60 & 59.11 \\
                 &                       & 50\%     & 62.14 & 71.98 & 48.91     & 53.35      & 55.13  & 32.68  & 37.00 & 51.60 \\
    \midrule
    \multirow{8}{*}{LLM-Pruner}
                 & \multirow{4}{*}{C4}   & 20\%     & 71.68 & 79.54 & 74.95     & 67.48      & 74.33  & 46.84  & 44.60 & 65.63 \\
                 &                       & 30\%     & 66.97 & 79.16 & 70.58     & 65.04      & 67.47  & 42.66  & 41.00 & 61.84 \\
                 &                       & 40\%     & 62.78 & 75.46 & 60.77     & 58.33      & 56.23  & 33.62  & 38.20 & 55.06 \\
                 &                       & 50\%     & 62.02 & 70.51 & 49.34     & 53.75      & 43.52  & 27.99  & 33.80 & 48.70 \\
    \cmidrule(l){2-11}
                 & \multirow{4}{*}{CMQA} & 20\%     & 66.51 & 74.92 & 58.47     & 62.67      & 66.41  & 35.92  & 36.20 & 57.30 \\
                 &                       & 30\%     & 61.77 & 67.19 & 36.64     & 51.54      & 51.01  & 25.26  & 30.40 & 46.26 \\
                 &                       & 40\%     & 58.93 & 61.43 & 31.23     & 49.25      & 42.51  & 22.78  & 27.80 & 41.99 \\
                 &                       & 50\%     & 61.96 & 58.11 & 28.81     & 51.70      & 36.78  & 22.27  & 26.80 & 40.92 \\
    \midrule
    \multirow{8}{*}{FLAP}
                 & \multirow{4}{*}{C4}   & 20\%     & 70.89 & 80.20 & 77.62     & 70.80      & 72.18  & 46.59  & 44.80 & 66.15 \\
                 &                       & 30\%     & 70.37 & 79.43 & 75.05     & 68.43      & 68.27  & 44.11  & 43.80 & 64.21 \\
                 &                       & 40\%     & 67.00 & 77.20 & 70.60     & 67.09      & 65.74  & 43.09  & 41.40 & 61.73 \\
                 &                       & 50\%     & 62.75 & 73.56 & 63.09     & 62.27      & 57.53  & 39.42  & 37.60 & 56.60 \\
    \cmidrule(l){2-11}
                 & \multirow{4}{*}{CMQA} & 20\%     & 73.00 & 79.92 & 72.26     & 66.85      & 73.74  & 45.82  & 43.40 & 65.00 \\
                 &                       & 30\%     & 69.94 & 78.35 & 67.79     & 65.11      & 71.72  & 44.88  & 45.20 & 63.28 \\
                 &                       & 40\%     & 66.21 & 76.82 & 64.74     & 63.14      & 65.53  & 42.58  & 40.20 & 59.89 \\
                 &                       & 50\%     & 64.16 & 74.48 & 57.94     & 57.77      & 60.82  & 38.65  & 40.20 & 56.29 \\
    \midrule
    \multirow{8}{*}{ShortGPT}
                 & \multirow{4}{*}{C4}   & 20\%     & 61.80 & 74.16 & 70.62     & 70.17      & 65.87  & 42.49  & 40.80 & 60.84 \\
                 &                       & 30\%     & 37.77 & 69.75 & 57.88     & 69.30      & 52.90  & 35.84  & 38.40 & 51.69 \\
                 &                       & 40\%     & 62.20 & 61.81 & 47.23     & 62.51      & 44.87  & 31.83  & 35.60 & 49.43 \\
                 &                       & 50\%     & 62.20 & 58.43 & 40.87     & 61.40      & 37.21  & 31.57  & 30.80 & 46.07 \\
    \cmidrule(l){2-11}
                 & \multirow{4}{*}{CMQA} & 20\%     & 61.80 & 74.16 & 70.62     & 70.17      & 65.87  & 42.49  & 40.80 & 60.84 \\
                 &                       & 30\%     & 61.53 & 69.80 & 64.57     & 69.53      & 55.47  & 39.33  & 37.80 & 56.86 \\
                 &                       & 40\%     & 44.98 & 65.02 & 33.19     & 51.66      & 66.46  & 46.17  & 33.60 & 48.73 \\
                 &                       & 50\%     & 62.17 & 52.61 & 33.35     & 56.91      & 31.73  & 26.71  & 28.80 & 41.75 \\
    \midrule
    \multirow{8}{*}{OWL}
                 & \multirow{4}{*}{C4}                     & 20\%     & 75.44  & 79.22  & 77.79     & 71.82      & 71.80 & 45.05 & 45.20  & 66.62  \\
                 & & 30\%     & 69.91  & 78.84  & 75.36     & 68.19      & 68.43 & 43.34 & 42.80  & 63.84  \\
                 & & 40\%     & 66.39  & 76.77  & 69.58     & 63.38      & 62.58 & 39.76 & 40.60  & 59.87  \\
                 & & 50\%     & 63.49  & 72.69  & 59.25     & 58.48      & 49.20 & 31.83 & 38.80  & 53.39  \\
    \cmidrule(l){2-11}
                 & \multirow{4}{*}{CMQA} 
                 & 20\%     & 76.48  & 80.30  & 77.19     & 68.35      & 74.71 & 46.84 & 45.00  & 66.98  \\
                 & & 30\%     & 69.82  & 78.40  & 73.28     & 64.09      & 71.76 & 43.17 & 42.40  & 63.27  \\
                &  & 40\%     & 69.97  & 77.69  & 66.97     & 60.62      & 67.51 & 42.92 & 40.20  & 60.84  \\
                &  & 50\%     & 63.46  & 73.18  & 55.15     & 54.78      & 57.58 & 37.29 & 37.80  & 54.17  \\
                 
    \midrule
    \multirow{12}{*}{GISP}
                 & \multirow{4}{*}{C4 + Perplexity}           & 20\%     & 73.30 & 78.45 & 73.09     & 68.11      & 67.59  & 42.92  & 42.00 & 63.64 \\
                 &                              & 30\%     & 69.20 & 76.71 & 69.68     & 65.98      & 62.67  & 37.46  & 40.80 & 60.36 \\
                 &                              & 40\%     & 65.14 & 73.67 & 62.80     & 61.64      & 54.55  & 33.87  & 36.80 & 55.49 \\
                 &                              & 50\%     & 58.41 & 68.28 & 50.79     & 58.64      & 44.02  & 28.41  & 32.20 & 48.68 \\
    \cmidrule(l){2-11}
                 & \multirow{4}{*}{CMQA + Perplexity}         & 20\%     & 80.80 & 77.86 & 76.39     & 71.82      & 74.75  & 46.33  & 42.20 & 67.16 \\
                 &                              & 30\%     & 80.83 & 75.52 & 71.91     & 69.69      & 72.22  & 44.71  & 41.60 & 65.21 \\
                 &                              & 40\%     & 79.54 & 72.52 & 63.36     & 67.88      & 67.85  & 41.81  & 39.20 & 61.74 \\
                 &                              & 50\%     & 76.18 & 67.52 & 51.67     & 61.56      & 59.47  & 36.77  & 36.40 & 55.65 \\
    \cmidrule(l){2-11}
                 & \multirow{4}{*}{CMQA + Margin} & 20\% & 80.28 & 79.00 & 76.83 & 72.22      & 75.17  & 45.99  & 43.80 & 67.61 \\
                 &                              & 30\%     & 81.16 & 76.77 & 72.87     & 71.59      & 72.94  & 45.73  & 41.80 & 66.12 \\
                 &                              & 40\%     & 80.00 & 73.83 & 65.79     & 70.09      & 70.16  & 43.52  & 40.00 & 63.34 \\
                 &                              & 50\%     & 72.97 & 69.91 & 55.15     & 65.59      & 63.09  & 38.23  & 37.60 & 57.50 \\
    \bottomrule
  \end{tabular}
  }
\end{table*}

\end{document}